\RequirePackage[l2tabu]{nag}
\documentclass{article}

\usepackage[utf8]{inputenc}
\usepackage[T1]{fontenc}

\usepackage{geometry}
\usepackage{color}

\usepackage{lmodern}
\usepackage[tracking=true,kerning=true,final]{microtype}

\usepackage{hyperref}
\hypersetup{colorlinks=true, linkcolor=blue,  anchorcolor=blue, citecolor=blue, filecolor=blue, menucolor=blue, urlcolor=blue}
\hypersetup{pdfauthor={},pdftitle={}}

\usepackage{caption}   
\usepackage{subcaption}
\usepackage{color}
\usepackage{graphicx}
\graphicspath{{./}{./figs/}}

\usepackage{amsfonts}
\usepackage{bm}
\usepackage{mathtools}

\DeclareMathOperator{\var}{\text{Var}}

\title{Output-weighted optimal sampling for Bayesian  regression and rare event statistics using few samples}

\author{Themistoklis P. Sapsis
\thanks{Corresponding author: \href{mailto:sapsis@mit.edu}{sapsis@mit.edu},
Tel: (617) 324-7508, Fax: (617) 253-8689%
}\\
Department of Mechanical Engineering,
\\ Massachusetts Institute of Technology, \\
77 Massachusetts Ave., Cambridge, MA 02139}
\date{\today}

\usepackage{amsmath}
\usepackage{amssymb}

\usepackage[normalem]{ulem}

\begin{document}

 \maketitle\ 
\begin{abstract}

For many important problems the quantity of interest is an unknown
function of the parameters, which is a random vector with
known statistics. Since the dependence of the output on this random vector
is unknown, the challenge is to identify its statistics, using the minimum
number of function evaluations. This problem can been seen in the
 context of active learning or optimal experimental design. We employ Bayesian
regression to represent the derived model uncertainty due to finite and small
number of input-output pairs. In this context we evaluate existing methods
for optimal sample selection, such as model error minimization and mutual
information maximization.  {We show that for the case of known
output variance, the commonly employed criteria in the literature do not
take into account the output values of the existing input-output pairs, while
for the case of unknown output variance this dependence can be very weak}.
We introduce a criterion that takes into account the values of the output for the existing samples
and adaptively selects inputs from regions of the parameter
space which have important contribution to the output. The new method allows
for application to high-dimensional inputs, paving the way for
optimal
experimental design in high-dimensions. 

\textbf{Keywords}: Optimal experimental design; Rare extreme events; Bayesian regression; Optimal sampling; Active learning
  
\end{abstract}

\section{Introduction}
For a wide range of problems in engineering and science it is essential to quantify the statistics of specific quantities of interest (or output) that depend on uncertain parameters (or input) with known statistical characteristics. The main obstacle towards this goal is that this dependence is not known a priori and numerical or physical experiments need to be performed in order to specify it.  If the problem at hand allows for the generation of many input-output pairs then one can employ standard regression methods to machine learn the input-output map over the support of the input parameters and subsequently compute the statistics of the   output. 

However, for several problems of interest it is not possible to simulate even a moderate size of input parameters. In this case it is critical to choose the input samples carefully so that they provide the best possible information for the output of interest \cite{Chaloner95, marzouk12, Uhler19}. A class of problems that belong in this family is the probabilistic quantification of extreme or rare events rising from high dimensional complex systems such as turbulence \cite{qi15, Farazmande1701533, sapsis2018, Blonigan19, brunton2019a}, networks \cite{Sarkar2018}, waves \cite{cousinsSapsis2015_JFM, mohamad2016b, Majda18ex}, and materials or structures \cite{Ortiz05, fan19}. Of course the considered setup is not limited to extreme or rare events but it is also relevant for any problem where the aim is to quantify the input-output relationship with very few but carefully selected data points. 

The described setup is a typical example of an optimal experimental design or active learning problem \cite{Chaloner95}. Specifically, we will assume that we have already a sequence of input-output data and our goal will be to sequentially identify the next most informative input or experiment that one should perform in order to achieve fastest possible convergence for the output statistics. The problem has been studied extensively using criteria relying on mutual information theory or  the Kullback–Leibler divergence (see e.g. \cite{bilionis19}). More recently another criterion was introduced focusing on the rapid convergence of the output statistics \cite{mohamad2018}. A common characteristic of these methods is the large computational cost associated with the resulted optimization problem that constrains applicability to low-dimensional input or parameter spaces. 

The first objective of this work is to understand some fundamental limitations of popular selection criteria widely used for optimal experimental design (beyond the large computational cost).  Specifically, we will examine how well these criteria distinguish and promote  the input parameters that have the most important influence to the quantities of interest. The second objective is the formulation of a new, output-weighted  selection approach that
explicitly and in a controllable manner takes into account, beyond the uncertainty of each input parameter, its effect on the output variables, i.e. the quantities of interest. This is an important characteristic as it is often the case that a small number of input parameters controls a specific quantity of interest. The philosophy of the developed criterion is to exploit the existing samples in order to estimate which input parameters are the most influential for the input and then bias the sampling process using this information. 

Beyond its intuitive and controllable character on selecting input samples according to their effect to the output statistics, the new criterion has a numerically tractable form which allows for easy computation of each value and gradient. The latter property allows for the employment of gradient optimization methods and therefore the applicability of the approach even in high-dimensional input spaces. We demonstrate ideas through several examples ranging from linear to nonlinear maps with low and high dimensional input spaces. In particular, we show that the important dependencies of given quantities of interest can be identified and quantified using a very small number of input-output pairs, allowing also for quantification of rare event statistics with minimal computational cost. 
\section{Setup}
Let the input vector $\mathbf{x} \in \mathbb{R}^m $
denote the set of parameters or system variables and $\mathbf{y} \in \mathbb{R}^d$ be the output vector describing the quantities of interest. The input vector can be thought as high-dimensional with known statistics described by the probability density function (pdf) $p(\mathbf x)$ that corresponds to mean value $\mu_x$ and covariance $\mathbf{C}_{xx}$ (or correlation $\mathbf{R}_{xx}$). In what follows we will use $p$ to denote pdf and an index will be used only if the random variable is not automatically implied by the argument. 

A map  from the input to the output variables, $\mathbf{y}=\mathbf{T(x)}$,  exists and our aim is to approximate the statistics of the output, $p(\mathbf{y})$,  using the smallest possible number of evaluations of the map $\mathbf{T}$. We will assume that we have already  obtained some input-output pairs which we employ
in order to optimize the selection of the next input that one should evaluate. This problem can be seen as an optimal experimental design problem where the experimental parameters that one is optimizing coincide with the random input parameters. All the results/methods presented in this work can be formulated in the standard setup of optimal experimental design in a straightforward way.

The first step of the approach is to employ a Bayesian regression model to represent the map $\mathbf{T}$. Our choice of the Bayesian framework is dictated by our need to have a priori estimates for the model error, as those will be employed  for the sample selection criteria. For simplicity we will present our results for linear regression models, although the extension for regression schemes with nonlinear basis functions or Gaussian process regression schemes is straightforward. We formulate a linear regression model with an input vector $\mathbf{x}$ that multiplies a coefficient matrix $\mathbf{A}\in \mathbb{R}^{d\times m}$ to produce an output vector $\mathbf{y}$, with Gaussian noise added:
\begin{align}
\begin{split}
\mathbf{y}\ & = \mathbf{Ax}+\mathbf{e,}\\
\mathbf{e} & \sim\ \mathcal{N}(0,\mathbf{V}), \\
p(y|\mathbf{x},\mathbf{A},\mathbf{V}) & = \mathcal{N}(\mathbf{Ax},\mathbf{V}).
\end{split}
\end{align}
{
We emphasize that for what follows we consider the case of a known noise variance $\mathbf{V}\in \mathbb{R}^{d\times
d}$. The case of unknown
covariance matrix $\mathbf{V}$ is discussed in Appendix D.} The basic setup involves a given data set of pairs $D=\left\{ (\mathbf{y}_1,\mathbf{x}_1), (\mathbf{y}_2,\mathbf{x}_2),..., (\mathbf{y}_N,\mathbf{x}_N) \right\}$. We set $\mathbf{Y}=[\mathbf{y}_1,\mathbf{y}_2,...,\mathbf{y}_N]$ and $\mathbf{X}=[\mathbf{x}_1,\mathbf{x}_2,...,\mathbf{x}_N]$.

For the matrix $\mathbf{A}$ we assume a Gaussian prior with mean $\mathbf{M}\in \mathbb{R}^{d\times
m}$ and  covariance $\mathbf{K}\in \mathbb{R}^{m\times
m}$ for the columns, and $\mathbf{V}$ for the rows. This has the form:\begin{align}
p(\mathbf{A}) & \sim\mathcal{N}(\mathbf{M},\mathbf{V},\mathbf{K}) \ =\frac{|\mathbf{K}|^{d/2}}{|2\pi\mathbf{V}|^{m/2}}\exp(-\frac{1}{2} \text{tr}((\mathbf{A}-\mathbf{M})^T\mathbf{V}^{-1}(\mathbf{A-M})\mathbf{K})).
\end{align}
Then one can obtain  the posterior for the matrix $\mathbf{A}$ \cite{rasmu05, Thomas10}
\begin{align}\label{coef_dis}
p(\mathbf{A}|D,\mathbf{V}) & \sim\mathcal{N}(\mathbf{S}_{yx}\mathbf{S}_{xx}^{-1},\mathbf{V},\mathbf{S}_{xx}),
\end{align}
where,
\begin{align}
\begin{split}
\mathbf{S}_{xx} & =\mathbf{X}\mathbf{X}^T+\mathbf{K,} \label{eq:prior} \\
\mathbf{S}_{yx} & =\mathbf{Y}\mathbf{X}^T+\mathbf{MK.}
\end{split}
\end{align}
Essentially, $\mathbf{X}\mathbf{X}^T$ is the data correlation of  the  sample input points $\mathbf{x}_i, \ i=1,...,N$. We choose $\mathbf{K}=\alpha\mathbf{I}$ ($\mathbf{I}$ is the identity matrix) and $\mathbf{M}=0 $, where $\alpha$ is an empirical parameter that will be optimized.
 Therefore, the above relations take the form
\begin{align*}
\mathbf{S}_{xx} & =\mathbf{X}\mathbf{X}^T +\alpha \mathbf{I,}\\
\mathbf{S}_{yx} & =\mathbf{Y}\mathbf{X}^T.
\end{align*}
Based on the above we obtain the probability distribution function for new inputs $\mathbf{x}$:
\begin{align}\label{model1}
\begin{split}
p(\mathbf{y}|\mathbf{x},D,\mathbf{V}) & = \mathcal{N}(\mathbf{S}_{yx}\mathbf{S}_{xx}^{-1}\mathbf{x},\mathbf{V(}1+c)_{}), \\c & = \mathbf{x}^T\mathbf{S}_{xx}^{-1}\mathbf{x}.
\end{split}
\end{align}
Then we can obtain an estimate for the probability density function of the output as\begin{equation}
p(\mathbf{y|D,\mathbf{V}})=\int p(\mathbf{y}|\mathbf{x},D,\mathbf{V})p(\mathbf x) d\mathbf x.
\end{equation}It is important to emphasize that the output $\mathbf{y}$ is random due to two sources: i) the uncertainty of the input vector $\mathbf{x}$, and ii) the uncertainty due to the model error expressed by the term $c$. The latter is directly related to the choice of data-points $\mathbf{x}_i, \ i=1,...,N$ and the goal is to choose these points in such a way so that the statistics of $\mathbf{y}$ converge most rapidly.

The most notable property for this model is the fact that the model error is independent of the expected output value of the system. This fact holds true also for Gaussian Process regression (GPR) schemes or regression models that use nonlinear basis functions. This property will have very important consequences when it comes to the optimal input sample selection for the modeling of the input-output relation.

\subsection{Properties of the data correlation $\mathbf{S}_{xx}$}\label{propertyS}

One can compute the eigenvectors,  $ \mathbf{\hat r}_i$, and eigenvalues, $\sigma_i^2$, of the data correlation matrix,  $\mathbf{S}_{xx}$, \begin{equation*}
\mathbf{\hat R}=[\mathbf{\hat r}_1|...|\mathbf{\hat r}_m]
\in \mathbb{R}^{m\times m} \text{ \ and \ } \sigma_i^2, \ i=1,...,m.
\end{equation*}
By applying a linear transformation to the $\mathbf{S}_{xx}$ eigendirections, $\mathcal{X}=\mathbf{\hat R}^T\mathbf{ x}$ we have\begin{equation*}
\mathbf{x}^T\mathbf{S}_{xx}^{-1}\mathbf{x}=\sum_{i=1}^{m}\frac{\chi_i^2}{\sigma_i^2}.
\end{equation*}Thus, the eigendirections of $\mathbf{S}_{xx}$ indicate the principal directions of maximum confidence for the linear model. The eigenvalues quantify this confidence: the larger the eigenvalue the slower the uncertainty increases (quadratically) as $\chi_i^2$ increases.
For a new, arbitrary point, $\mathbf{x}_{N+1}=\mathbf{h}$ ,
 added to the family of 
$\mathbf{x}$ points we will have $\mathbf{X}'=[\mathbf{X}|\mathbf{x}_{N+1}]$.
By direct computation we obtain\begin{equation}\label{eq16}
\mathbf{S}'_{xx}=\sum_{i=1}^N\mathbf{x}_{i} \mathbf{x}^T_{i}=\mathbf{S}_{xx}+
 \mathbf{h} \mathbf{h}^T.
\end{equation}If the new point belongs to the $j$ eigendirection, $\mathbf{x}_{N+1}=\kappa \mathbf{r}_j$, where $\kappa \in \mathbb{R}$ then the new data correlation will be,\begin{equation*}
\mathbf{S}'_{xx}=\mathbf{S}_{xx}+\kappa^2  \mathbf{r}_{j} \mathbf{r}_{j}^T.
\end{equation*}     
It can be easily checked that under this assumption the new matrix $\mathbf{S}'_{xx}$ will have the same eigenvectors. Moreover the $j$ eigenvalue will be ${\sigma'}_j^2=\sigma_j^2+\kappa^2$, while all other eigenvalues will remain invariant. Therefore, by adding one more data-point along a principal direction will increase the confidence along this direction by the magnitude of this new point.

{The larger the magnitude of any point we add, the larger its impact on the covariance. One can trivially increase the magnitude of the new points but this does not offer any real benefit. Moreover, in a typical realistic scenario there will be magnitude constraints. To avoid this ambiguity, typical of linear regression problems, we will fix the magnitude of the input points, i.e. $\mathbf{x} \in \mathbb{S}^{m-1}=\{\mathbf{x}\in \mathbb{R}^m,\left\Vert \mathbf{x} \right\Vert=1\}$, so that we can assess the direction of the new points, without being influenced by the magnitude.} For nonlinear problems the input points should be chosen from a compact set, typically defined by the mechanics of the specific problem.  

\section{Fundamental limitations of standard optimal experimental design criteria}

Here we consider two popular criteria that can be employed for the selection of the next most informative input sample $\mathbf{x}_{N+1}$. The first one is based on the minimization of the model error expressed by the parameter $c$ (eq. (\ref{model1})), while the second one is the Kullback–Leibler divergence or equivalently the maximization of the mutual information between input and output variables, which is the standard approach in the optimal experimental design literature \cite{Chaloner95}.

We hypothesize a new input point $\mathbf{x}_{N+1}=\mathbf{h}$. As the corresponding output is not a priori known we will assume that it is given by the mean regression model, $\mathbf{y}_{N+1}=\mathbf{S}_{yx}\mathbf{S}_{xx}^{-1}\mathbf{x}_{N+1}$. The new pairs of data points will be $D'=\{D,(\mathbf{x}_{N+1},\mathbf{y}_{N+1})\}$. Under this setup the new model error
will be given by $c(\mathbf{x};\mathbf{h})=\mathbf{x}^T\mathbf{S}_{xx}'^{-1}\mathbf{x}$, where the new data correlation matrix is given by (\ref{eq16}). In addition, the mean estimate of the new model will remain invariant since,
\begin{align}
\label{mean_inv}
\begin{split}
\mathbf{S}'_{yx}\mathbf{S}_{xx}'^{-1}& =[\mathbf{S}_{yx}+\mathbf{S}_{yx}\mathbf{S}_{xx}^{-1}\mathbf{hh}^T][\mathbf{S}_{xx}+\mathbf{hh}^T]^{-1}\\
& = \mathbf{S}_{yx}\mathbf{S}_{xx}^{-1}[\mathbf{S}_{xx}+\mathbf{hh}^T][\mathbf{S}_{xx}+\mathbf{hh}^T]^{-1}\\
& = \mathbf{S}_{yx}\mathbf{S}_{xx}^{-1}.
\end{split}
\end{align}
\subsection{Minimization of the mean model error}
The first approach we will employ is to select $\mathbf{h}$ by minimizing the mean value of the uncertainty parameter 
$c$ (eq. (\ref{model1})).   
Using standard expressions for quadratic forms of a random variable \cite{Rencher08} we obtain  a closed expression, valid for any input distribution. More specifically, we will have:
\begin{align}\label{eq_mc}
\mu_{c}& =\mathbb{E}[\mathbf{x}^T\mathbf{S}_{xx}'^{-1}\mathbf{x}] =\text{tr}[\mathbf{S}'^{-1}_{xx}\mathbf{C}_{xx}]+\mu_x^T\mathbf{S}'^{-1}_{xx}\mu_x=\text{tr}[\mathbf{S}_{xx}'^{-1}\mathbf{R}_{xx}].
\end{align}
Moreover for the case of Gaussian input we also obtain \cite{Rencher08}
\begin{align}\label{eq_sc}
\sigma_{c}^2& =\var[\mathbf{x}^T\mathbf{S}_{xx}'^{-1}\mathbf{x}]  = 2\text{tr}[\mathbf{S}'^{-1}_{xx}\mathbf{C}_{xx}\mathbf{S}'^{-1}_{xx}\mathbf{C}_{xx}]+4\mu_x^T\mathbf{S}'^{-1}_{xx}\mathbf{C}_{xx}\mathbf{S}'^{-1}_{xx}\mu_x.
\end{align} 
 We note that the model uncertainty depends only on the statistics of the input $\mathbf{x}$ (expressed through the covariance $\mathbf{C}_{xx}$) and the samples $\mathbf{X}$ (expressed through the constant (i.e. non-dependent on $\mathbf x$) matrix $\mathbf{S}'_{xx}$). {In other words, the matrix $\mathbf{Y}$ and the output distribution play no role on the mean model uncertainty.
}

To understand the mechanics of selecting input samples using the mean model error we assume that $\mathbf{R}_{xx}$ is diagonal with eigenvalues $\sigma_i^2+\mu_{ x_i}^2
\ \ i=1,...,d,$ arranged with increasing order. We also assume that samples are collected only along the principal directions
of the input covariance. In this case the quantity that is minimized takes the form\begin{equation*}
\mu_c(\mathbf{h})=\text{tr}[\mathbf{S}_{xx}'^{-1}\mathbf{R}_{xx}]=\sum_{i=1}^m\frac{\sigma_i^2+\mu_{
x_i}^2}{n_i+\delta_{ik}}, \ \ \ {h}_i=\delta_{ik} \in \mathbb{S}^{m-1},
\end{equation*}
where $n_i$ denotes the number of samples in the $i^{th}$ direction. One should choose $\mathbf{h,}$ or equivalently $k,$ according to value of the derivative of $\mu_c(\mathbf{h})$. In particular,
\begin{equation*}
{h}_i=\delta_{ik}, \ \ k=\arg\min_i \left(-\frac{\sigma_i^2+\mu_{
x_i}^2}{n_i^2} \right).
\end{equation*} If all directions have been sampled with equal number (e.g. each of the directions have $n_i=1$), sampling will continue with the most uncertain direction. After sufficient sampling in this direction, the addition of a 
new sample will cause smaller effect than sampling the next most important
direction and this is when the scheme will change sampling direction. This
behavior guarantees that the scheme will never get `trapped' in one direction.
It will continuously evolve, as more samples in one direction lead to very
small eigenvalue of $\mathbf{S}_{xx}'^{-1}\mathbf{R}_{xx}$ along this direction,
and therefore sampling along another input direction will cause bigger contribution
to the trace.  

It is clear that sampling based on the uncertainty parameter $c$ searches
only in $\mathbf{x}-$directions with important uncertainty, while the impact of each input variable is completely neglected. \textit{Therefore, even directions that have zero effect on the output variable will still be sampled as long as they are uncertain.} 

\subsection{Maximization of the mutual information}\label{entr_conv}
An alternative approach for the
selection of a new sample, $\mathbf{x}_{N+1}=\mathbf{h} \in \mathbb{S}^{m-1}$,
is maximizing the entropy transfer or mutual information between the input
and output variables, when a new sample is added \cite{Chaloner95}:\begin{equation}
\mathcal{I}(\mathbf{x,y}|D',\mathbf{V}) =\mathcal{E}_{x}+\mathcal{E}_{y|D'}-\mathcal{E}_{x,y|D'}.
\end{equation} where each of the entropies above are defined as, 

\begin{align*}
\mathcal{E}_{x}  = \int p(\mathbf{x}) \log
p(\mathbf{x}) d \mathbf{x}, \  \ \ \ \ \   \mathcal{E}_{y|D'} = \iint p(\mathbf{y}|D',\mathbf{V}) \log
p(\mathbf{y}|D',\mathbf{V})d \mathbf{y}, 
\end{align*} 
\vspace{-5mm} 
\begin{align*}
\mathcal{E}_{x,y|D'} = \iint p(\mathbf{y,x}|D',\mathbf{V}) \log
p(\mathbf{y,x}|D',\mathbf{V}) d \mathbf{x} d \mathbf{y.}
\end{align*}
This is also equivalent with maximizing the mean value of the  Kullback–Leibler (KL) divergence \cite{marzouk12} \begin{align*}
\mathbb{E}^y[D_{KL}[p(\mathbf{x}|\mathbf{y},D')||p(\mathbf{x})]]&=\int_y\int_x p(\mathbf{x}|\mathbf{y},D',\mathbf{V})\log\frac{p(\mathbf{x}|\mathbf{y},D',\mathbf{V})}{p(\mathbf{x})}d\mathbf{x}p(\mathbf{y}|D',\mathbf{V})d\mathbf{y}\\
&= \int_y\int_x
p(\mathbf{x},\mathbf{y}|D',\mathbf{V})\log\frac{p(\mathbf{x},\mathbf{y}|D',\mathbf{V})}{p(\mathbf{x})p(\mathbf{y}|D',\mathbf{V})}d\mathbf{x} d\mathbf{y}\\
&=\mathcal{I}(\mathbf{x,y}|D',\mathbf{V}).
\end{align*} 
We first compute the entropy    of $p(\mathbf{x,y}|D',\mathbf{V})$:\begin{align*}\label{entr2}
\mathcal{E}_{x,y}(\mathbf{h}) & = \iint p(\mathbf{y,x}|D',\mathbf{V}) \log
p(\mathbf{y,x}|D',\mathbf{V}) d \mathbf{x} d \mathbf{y} \\
& = \iint p(\mathbf{y}|\mathbf{x},D',\mathbf{V})p(\mathbf{x})
\log p(\mathbf{y}|\mathbf{x}, D',\mathbf{V}) d \mathbf{x} d \mathbf{y}+\iint
p(\mathbf{y}|\mathbf{x},D',\mathbf{V})p(\mathbf{x})
\log p(\mathbf{x}) d \mathbf{x} d \mathbf{y}\\
& = \int \mathcal{E}_{y|x}(\mathbf{x;h})\ p(\mathbf{x})
 d \mathbf{x} +\int
p(\mathbf{x})
\log p(\mathbf{x}) d \mathbf{x} \\
& = \mathbb{E}^{x}[\mathcal{E}_{y|x}(\mathbf{x;h})]+\mathcal{E}_{x}.
\end{align*}
We focus on computing the first term on the right hand side. For the linear
regression model, the conditional error follows a Gaussian distribution.
From standard expressions about the entropy of a multivariate Gaussian we
have\begin{align*}
\mathcal{E}_{y|x}(\mathbf{x;h})=\frac{1}{2}\log (1+c)^d|2\pi e\mathbf{V}|=\frac{d}{2}\log
(1+c(\mathbf{x;h}))+\frac{1}{2}\log|2\pi e\mathbf{V}|.
\end{align*}
Therefore, \begin{equation*}\label{entr3}
\mathbb{E}^{x}[\mathcal{E}_{y|x}(\mathbf{x;h})]=\frac{d}{2}\mathbb{E}^{x}[\log(1+c(\mathbf{x;h}))]+\frac{1}{2}\log|2\pi
e\mathbf{V}|.
\end{equation*}
In the general case, we cannot compute the entropy of the output, conditional
on $D'$. To this end, the mutual information of the input and output, conditioned
on $D',$ takes the form\begin{equation}\label{info_full}
\mathcal{I}(\mathbf{x,y}|D',\mathbf{V}) =\mathcal{E}_{y}(\mathbf{h})-\frac{d}{2}\mathbb{E}^{x}[\log(1+c(\mathbf{x;h}))]-\frac{1}{2}\log|2\pi
e\mathbf{V}|.
\end{equation}
This expression is valid for any input distribution and relies only on the
assumption of Bayesian linear regression. To compute the involved terms,
one has to perform a Monte-Carlo or importance sampling approach, even for linear regression models and Gaussian inputs. This, of course, limits the applicability of the approach to very low-dimensional input spaces. {We note that the above expression is valid for the case of known noise variance, $\mathbf{V}$. The case of unknown  variance $\mathbf{V}$ is considered in Appendix D.}
\subsubsection{Gaussian approximation of the output}
 To overcome this computational obstacle one can consider an analytical approximation
of the mutual entropy, assuming Gaussian statistics for the output. This
assumption is not true in general, even for Gaussian input, because of the multiplication
of the (Gaussian) uncertain model parameters (matrix $\mathbf{A}$) with the
the Gaussian input (vector $\mathbf{x}$).

We focus on the computation of the entropy of the output $\mathbf{y}$, so
that we can derive an expression for the mutual information. We will approximate
the pdf for $\mathbf{y}$ through its second order statistics. Given that
the input variable is Gaussian and the exact model is linear the Gaussian
approximation for the output is asymptotically accurate. Still, it will help
us to obtain an understanding of how the criterion works to select new samples.

We express the covariance of the output variable using the law of total variance
\begin{align}\label{cov_spl}
\mathbf{C}_{{yy}}(D',\mathbf{V})& =\mathbb{E}^x[\mathbf{C}_{yy|x}(D',\mathbf{V})]+\text{cov}[\mathbb{E}^y(\mathbf{y|x},D',\mathbf{V})].
\end{align}
The first term is the average of the updated conditional covariance of the
output variables and it is capturing the regression error. The second term
expresses the covariance due to the uncertainty of the input variable $\mathbf{x}$,
as measured by the estimated regression model using the input data in $D'$.

As we pointed out earlier the mean model using either $D$ or $D'$ remains invariant. Therefore, we have  
\begin{equation*}
 \mathbf{C}_{{yy}}(D',\mathbf{V})=\mathbf{V}(1+\mu_c(\mathbf{h}))+\mathbf{S}_{yx}\mathbf{S}_{xx}^{-1}\mathbf{C}_{xx}\mathbf{S}_{xx}^{-1}\mathbf{S}^T_{yx}.
\end{equation*}
In this way we have the approximated entropy of the output variable using a
Gaussian approximation, which is also an upper bound for any other non-Gaussian
distribution with the same second order statistics
\begin{align*}
\mathcal{E}_{y}(\mathbf{h}) & = \frac{1}{2}\log|\mathbf{V}(1+\mu_c(\mathbf{h}))+\mathbf{S}_{yx}\mathbf{S}_{xx}^{-1}\mathbf{C}_{xx}\mathbf{S}_{xx}^{-1}\mathbf{S}^T_{yx}|+\frac{d}{2}\log(2\pi
e).
\end{align*}
Therefore, we have the second-order statistics approximation of the mutual
information in terms of the new sample $\mathbf{h} \in \mathbb{S}^{m-1}$,
denoted as $\mathcal{I}_G$: \begin{align}
\label{mutualc}
\begin{split}
\mathcal{I}_G(\mathbf{x,y}|D',\mathbf{V}) & =\frac{1}{2}\log|\mathbf{V}(1+\mu_c(\mathbf{h}))+\mathbf{S}_{yx}\mathbf{S}_{xx}^{-1}\mathbf{C}_{xx}\mathbf{S}_{xx}^{-1}\mathbf{S}^T_{yx}|-\frac{1}{2}\log|\mathbf{V}|\\
& -\frac{d}{2}\mathbb{E}^{x}[\log(1+c(\mathbf{x;h}))].
\end{split}
\end{align}
We observe that the second-order approximation of the mutual information criterion has minimial dependence on the output samples $\mathbf{Y}$. Specifically,
(\ref{mutualc}) depends on the uncertainty parameter $c$ and its statistical moments, as well as the term $\mathbf{S}_{yx}\mathbf{S}_{xx}^{-1}\mathbf{C}_{xx}\mathbf{S}_{xx}^{-1}\mathbf{S}^T_{yx}$. However, the latter, is not coupled with the new hypothetical point $\mathbf{h}$ and to this end the minimization of this criterion does not guarantee that the output values will be taken into account in a meaningful way. Instead, the selection of the new sample, depends primarily on minimizing
$\mu_c=\text{tr}[\mathbf{S}_{xx}'^{-1}\mathbf{R}_{xx}]$, always under the constraint
 $\left\Vert \mathbf{h} \right\Vert=1$,
 a process that depends exclusively on the current samples $\mathbf{X}$ and
the statistics of the input $\mathbf{x}$. 

Therefore, regions of $\mathbf{x}$
associated with large or important values of the output $\mathbf{y}$ are
not emphasized by this sampling approach and the emphasis is given in regions
that minimize the mean model error $\mu_c$. We note that these conclusions are valid for the second-order
approximation of the mutual information criterion, { with known output variance, $\sigma_V^2$. When one considers the mutual information criterion with unknown output variance which has to be inferred using conjugate priors, there is dependence of the criterion on the output vector $\mathbf{Y}$. However, this dependence may be very weak or even zero depending on inference parameters that are optimized based on the data. See Appendix D for details.}

\subsection{Nonlinear basis regression}

Similar conclusions can be made for the case where one utilizes nonlinear
basis functions. In this case we assume that the input points `live' within a compact set$\mathcal{.}$ Specifically, let the input $\mathbf{x} \in \mathcal{X}\subset\mathbb{R}^m ,$
be expressed as a function of another input $\mathbf{z} \in \mathbb{\mathcal{Z}\subset R}^s$
where the input value has distribution $p(\mathbf{z})$ and $\mathcal{Z}$ be a compact set.   One can choose a
set of basis functions\begin{equation}
\mathbf{x}=\phi(\mathbf{z}).
\end{equation}
 In this case the distribution of the output values will be given by :
\begin{align}\label{model2}
\begin{split}
p(\mathbf{y}|\mathbf{z},D,\mathbf{V}) & = \mathcal{N}(\mathbf{S}_{y\phi}\mathbf{S}_{\phi\phi}^{-1}\mathbf{\phi(\mathbf{z})},\mathbf{V(}1+c)_{}),
\\c & = \mathbf{\phi{(\mathbf{z})}}^T\mathbf{S}_{\phi\phi}^{-1}\phi(\mathbf{z}).
\end{split}
\end{align}
The mean of the model uncertainty parameter $c=\phi(\mathbf{z})\mathbf{S}^{-1}_{\phi\phi}\phi(\mathbf{z})^T$ will become\begin{equation}
\mu_{c} =\text{tr}[\mathbf{S}^{-1}_{\phi\phi}\mathbf{C}_{\phi\phi}]+\mu_\phi^T\mathbf{S}^{-1}_{\phi\phi}\mu_\phi=\text{tr}[\mathbf{S}^{-1}_{\phi\phi}\mathbf{R}_{\phi\phi}],
\label{eq:mu_NL}\end{equation}
where \begin{equation*}
\mathbf{S}_{ \phi\phi}=\sum_{i=1}^N\mathbf{\phi}{(\mathbf{z}_i)} \mathbf{\phi}{(\mathbf{z}_i)}^T \ \ \text{and} \ \ \ \mathbf{S}'_{ \phi\phi}=\mathbf{S}_{ \phi\phi}+\mathbf{\phi}{(\mathbf{h})} \mathbf{\phi}{(\mathbf{h})}^T,
\end{equation*}
and
\begin{equation*}\label{mu_cor_phi}
\mu_\phi=\int \phi(\mathbf{z})f_{z}(\mathbf{z})d\mathbf{z} \ \ \ \
\text{and} \ \ \ \ \mathbf{C}_{\phi\phi}= \int\left( \phi(\mathbf{z})-\mu_\phi \right)(\phi(\mathbf{z})-\mu_\phi)^T
f_{z}(\mathbf{z})d\mathbf{z}. 
\end{equation*}
Following the same steps as we did for the linear model, we will have, first for the conditional entropy (assuming that the model noise in the nonlinear case is Gaussian)\begin{align*}
\mathcal{E}_{y|z}(\mathbf{z;h})=\frac{1}{2}\log (1+c)^d|2\pi e\mathbf{V}|=\frac{d}{2}\log
(1+c(\phi(\mathbf{z});\mathbf{h}))+\frac{1}{2}\log|2\pi e\mathbf{V}|.
\end{align*}
Therefore, 
\begin{equation*}
\mathbb{E}^{z}[\mathcal{E}_{y|z}(\mathbf{z;h})]=\frac{d}{2}\mathbb{E}^z[\log
(1+c(\phi(\mathbf{z});\mathbf{h})]+\frac{1}{2}\log|2\pi
e\mathbf{V}|.
\end{equation*}
The exact expression for the mutual information for the nonlinear case will be:
\begin{align}
\mathcal{I}(\mathbf{x,y}|D',\mathbf{V})=\mathcal{E}_y-\frac{d}{2}\mathbb{E}^z[\log
(1+c(\phi(\mathbf{z});\mathbf{h})]-\frac{1}{2}\log|2\pi
e\mathbf{V}|.
\end{align}
To perform, the second-order statistical approximation for the entropy $\mathcal{E}_y$, we follow the same steps as for the linear model case to obtain\begin{align}
\begin{split}
\mathcal{I}_G(\mathbf{x,y}|D',\mathbf{V})
& =\frac{1}{2}\log|\mathbf{V}(1+\mu_c(\mathbf{h}))+\mathbf{S}_{y\phi}\mathbf{S}_{\phi\phi}^{-1}\mathbf{C}_{\phi\phi}\mathbf{S}_{\phi\phi}^{-1}\mathbf{S}^T_{y\phi}|-\frac{1}{2}\log|\mathbf{V}|\\
& -
\frac{d}{2}\mathbb{E}^z[\log
(1+c(\phi(\mathbf{z});\mathbf{h})].
\end{split}
\end{align}The sampling strategy is more complicated in this case due to the nonlinearity
of the basis elements. However, even in the present setup the sampling depends
exclusively on the statistics of the input variable $\mathbf{z}$ and the
form of the basis elements $\phi$. The measured output values of the modeled process
do not enter explicitly into the optimization procedure for the next sample, in the same fashion with the linear model.

\section{Optimal sample selection considering the output values}
\begin{figure}[t]
\begin{center}
\includegraphics[width=0.98\textwidth]{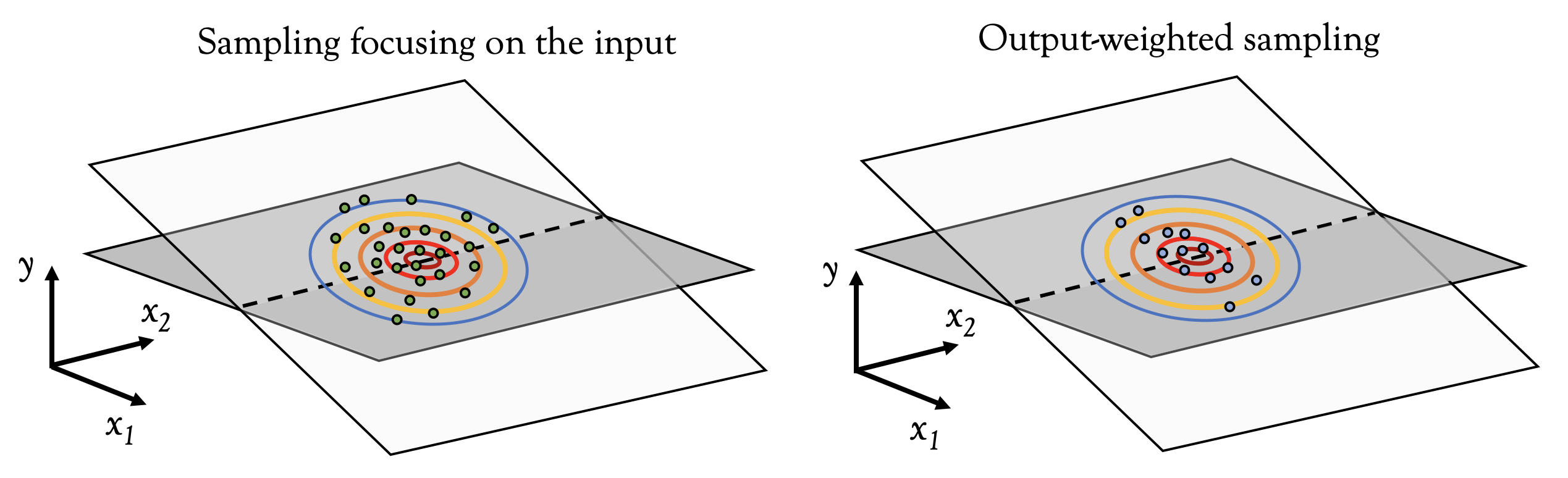}
\end{center}
\setlength{\abovecaptionskip}{-12pt}
\caption{ {A schematic of an input space (grey plane), the pdf of the input (colored contours), and the output surface (white plane). A criterion that only takes into account the characteristics of 
the input, $\mathbf{x,}$   focuses on sampling the input directions according
to their variance (left). However,  not all input dimensions have the
same effect to the output, $\mathbf{y}$. Here only $x_1$
contributes to the output.  By utilizing a regression  trained with the existing
samples we quantify the effect of each input direction to the output and
select more samples from the important input dimensions (right). This expedites the convergence of the  output statistics, $p_y$. }   }
\label{fig:framework00}
\end{figure}

{
We saw that selecting input samples based on either the  mean model error or the mutual information does not effectively take into account  the output values of the existing samples. Our goal is to develop an approach that i) will give  emphasis on the output
values of the existing samples, and ii) will be computationally tractable. 
In \cite{mohamad2018}
a similar problem was considered where the goal was to design a sampling method that will accelerate the convergence of the pdf in regions associated with rare events. In particular the following steps were followed in \cite{mohamad2018}:\begin{enumerate}
\item 
Using the existing samples the authors obtain an estimate of the map (denote it as $\mathbf{y}_0(\mathbf{x})$), as well as, the map-estimation-error, $\sigma_{y_0}^2(\mathbf{x})$, at every point $\mathbf{x}$.
\item
This map-estimate and its error are then used to estimate the output pdf, denoted as $p_{y_0}(\mathbf{y})$, as well as, the pdf of the perturbed map along the direction of the map-estimation-error, denoted as $p_{y_0^+}(\mathbf{y}).$ 
\item A new hypothetical sample point, $\mathbf{h}$, was assumed and its impact first on the map-estimation-error, $\sigma_{y_0}^2(\mathbf{x;\mathbf{h}}),$ and then on the pdf of the perturbed map, $p_{y_0^+}(\mathbf{y;h})$, was quantified.

\item
Then the goal was to select the new sample that minimizes the distance of the two pdfs, i.e. between $p_{y_0^+}(\mathbf{y;h})$ and $p_{y_0}(\mathbf{y})$. As a distance the authors considered the $L_1$ difference of the logarithms of the two pdfs, instead of the KL divergence. The reason was that in the KL divergence the difference of the logarithms is multiplied with the pdf itself and therefore rare events play a less important role on the value of the criterion. By considering only the difference of the logarithms gave more emphasis in the regions associated with rare events. 
\end{enumerate}
The approach was very effective on computing the rare event properties (tails of the pdf) for arbitrary quantities of interest with a very few samples. However, it was limited by the large cost related to the computation of the two pdfs mentioned above, which was performed with direct Monte-Carlo methods.
For this reason the method could be applied in problems with relatively low-input dimensionality.

  In the present work we are going to build on \cite{mohamad2018} to derive a new criterion that follows the same principles as the one just described but it is also computationally tractable and can be used beyond the context of rare events, i.e. for general optimal experimental design problems with a large number of parameters. 

Specifically, we are going to apply the following steps: 
\begin{enumerate}
\item 
We will employ an asymptotic form of the criterion in \cite{mohamad2018} that will provide, analytically, the distance of the pdf logarithms, i.e. the pdf of the map-estimate and the pdf of the asymptotically-perturbed map-estimate. \item Using standard inequalities for norms of derivatives we will bound the asymptotic form of the criterion by a more intuitive and tractable form. This new form has an interesting interpretation as it naturally weights the importance of the estimated map-error by the pdf of the input but also the inverse of the pdf of the output. In this way more emphasis is given to inputs associated with large values of the output.
\item 
The final step of our analysis is to demonstrate how the derived bound can be analytically approximated in terms of second-order properties of the input pdf, as well as, second-order properties of the estimated output. This last step is the most cumbersome but also crucial in order to apply the method in high-dimensional problems, as the analytical approximation of the criterion allows for the application of gradient optimization methods. 
\end{enumerate}

In {Figure
\ref{fig:framework00}} we provide a sketch of the main idea. A two dimensional input space is shown where the output is a function that primarily depends on one input variable. The input variable has important variance in both dimensions. While methods relying on mutual information give emphasis primarily on the statistics of the input variable, resulting an-equally-good approximation of the map over all input dimensions through a uniform coverage of all input dimensions, an output-weighted approach assess the important of each candidate input sample by its effect on the statistics of the output, i.e. the quantity of interest. In this way the output values for the observable of interest are taken into account explicitly and in a controllable manner.}

\subsubsection*{Derivation of the output-weighted criterion}

Our goal is to compute samples that accelerate the convergence of the output statistics, expressed by the probability density function, $p(\mathbf{y})$. To measure how well this convergence has occurred, we are going to rely on the distance between the probability density function of the mean model\begin{equation}
\mathbf{y}_0=\mathbf{S}_{yx}\mathbf{S}_{xx}^{-1}\mathbf{x,}
\end{equation}    
and the perturbed model along the most important direction of the model uncertainty (dominant eigenvector of $\mathbf{V}$), denoted as $\mathbf{r_V}$:\begin{equation}
\mathbf{y}_+=\mathbf{S}_{yx}\mathbf{S}_{xx}^{-1}\mathbf{x} +\beta
\mathbf{r}_V (1+\mathbf{x}^T\mathbf{S}_{xx}'^{-1}\mathbf{x}),
\end{equation}where $\beta\ll1$ is a small scaling factor.
The corresponding probability density functions, $p_{{y}_0}(\mathbf{y})$ and $p_{{y}_+}(\mathbf{y})$ will differ only due to errors of the Bayesian regression, which vary as $\mathbf{h}$ changes. It is therefore meaningful to select the next sample that will  minimize their distance.
Moreover, as we are interested on capturing the probability density function equally well in regions of low and large probability we will consider the difference between the logarithms. Specifically, we define,
\cite{mohamad2018},\begin{equation}
\label{norm11}
D_{Log^1}(\mathbf{y}_+\Vert\mathbf{y}_0;\mathbf{h})=\int_{S_{y}}|\log p_{{y}_+}(\mathbf{y})-\log p_{{y}_0}(\mathbf{y})|d\mathbf{y,} 
\end{equation} 
where $S_{y}$ is a finite domain over which we are interested to define the criterion. Note that the latter has to be finite in order to have a bounded value for this distance. It can be chosen so that it contains
several standard deviations of the output process. The defined criterion focuses exactly on our goal which is the convergence of the output statistics, while the logarithm guarantees that even low probability regions have converged as well. This criterion for selecting samples was first defined in \cite{mohamad2018} and was shown that it results in a very effective strategy for sampling processes related to low-probability extreme events.  However, it is also associated with a very expensive optimization problem that has to be solved in order to minimize this distance. Apart from the cost, its complicated form does not allow for the application of gradient methods for optimization and therefore it is practical only for low-dimensional input spaces where non-gradient methods can be applied.
{
Here one of our goals is to study its relationship with existing criteria. We are also aiming to bound it by a more trackable form that is applicable for gradient optimization methods.}  

{To study the relationship of the criterion (\ref{norm11}) with the KL divergence, we  note that for bounded probability density functions the following inequality holds} \begin{equation}
D_{KL}(\mathbf{y}_0\Vert\mathbf{y}_+;{S_ y})= \int_{S_{y}}\left(
\log p_{{y}_+}(\mathbf{y})-\log
p_{{y}_0}(\mathbf{y})\right)p_{{y}_0}(\mathbf{y})d\mathbf{y}\leqslant \kappa
D_{Log^1}(\mathbf{y}_+\Vert\mathbf{y}_0),
\end{equation}
where $\kappa$ is a constant. To this end, the criterion based on the difference
of the logarithms is more conservative (i.e. harder to minimize) compared
with the KL divergence (defined over the same domain).

{Our next goal is to bound the $D_{Log^1}$ criterion by one that is more tractable to optimize. We consider the criterion (\ref{norm11}) for an asymptotically small value of $\beta$. This form of the criterion essentially expresses the infinitesimal difference between the mean model and the infinitesimally perturbed model by $\beta {\sigma_{{y}}^2}$. To compute analytically the value of the criterion for $\beta\rightarrow0$  we employ an asymptotic result originally obtained in \cite{mohamad2018} for the study of the criterion for a large number of input samples, i.e. very small ${\sigma_{{y}}^2}$. For this case, or equivalently the case where $\beta$ is very small that we are interested here, we have
the asymptotic form (Theorem 1 in \cite{mohamad2018})}\begin{equation}\label{Dlog1}
D_{Log^1}(\mathbf{y}_+\Vert\mathbf{y}_0;\mathbf{h})\simeq \beta\int_{S_y}\frac{\left\vert\frac{d }{ds }\mathbb{E}[{\sigma_{{y}}^2}(\mathbf{x;\mathbf{h}})
\cdot \boldsymbol 1_{{{y}_0}({x})= \text{s}}]\right\vert }{p_{{y}_0}(\mathbf{s})}d\mathbf{s},
\end{equation}
where, 
\begin{equation*}
{\sigma_{{y}}^2}(\mathbf{x;h})= \text{tr}(\text{cov}[\mathbf{y}|\mathbf{x},D'])=\text{tr}(\mathbf{V})(1+c(\mathbf{x;h})),
\end{equation*} 
is the conditional variance (on $\mathbf{x}$) if the output is scalar or the trace of the output conditional covariance matrix in the general case, while $\mathbf{y}_0(\mathbf{x})$ is the mean model from the input-output data collected so far. 

Using standard inequalities for the derivatives of differential functions one can bound the derivative in (\ref{Dlog1}). Specifically, if  the function $\mathbb{E}[{\sigma^2_{{y}}}(\mathbf{x})
\cdot \boldsymbol 1_{{{y}_0}({x})= \text s}]$ has uniformly bounded second derivative (with respect to a hypothetical new point $\mathbf{h}$), and $p_{{y}_0}(\mathbf{s})$ has not zeros or singular points, there exists a constant $\kappa_0$  such that (\cite{Kwong92}, Theorem 3.13, p. 109)\begin{equation}\label{ineq10}
 \int_{S_y}\frac{\left\vert\frac{d
}{ds }\mathbb{E}[{\sigma_{{y}}^2}(\mathbf{x})
\cdot \boldsymbol 1_{{{y}_0}({x})= \text{s}}]\right\vert }{p_{{y}_0}(\mathbf{s})}d\mathbf{s}\leqslant \kappa_{0} \int_{S_y}\frac{\mathbb{E}[{\sigma_{{y}}^2}(\mathbf{x})
\cdot \boldsymbol 1_{{{y}_0}({x})= \text{s}}] }{p_{{y}_0}(\mathbf{s})}d\mathbf{s}.
\end{equation}
Moreover,\begin{align*}
\int_{S_y}\frac{\mathbb{E}[{\sigma_{y}^2}(\mathbf{x})
\cdot \boldsymbol 1_{{{y}_0}({x})= \text{s}}] }{p_{y_0}(\mathbf{s})}d\mathbf{s} & =\mathbb{E} \left[ \frac{{\sigma_y^2}(\mathbf{x})}{p_{y_0}({\mathbf{y}_0}(\mathbf{x}))} \middle | S_x \right] = \int_{S_x}\frac{p_{x}(\mathbf{x})
 }{p_{{y}_0}({\mathbf{y}_0}(\mathbf{x}))}{\sigma_{ y}^2}(\mathbf{x})d\mathbf{x},
\end{align*}
where $S_x$ is the inverse image of the domain $S_y$ through the map, $\mathbf{y}_0(\mathbf{x})$. Based on this we obtain the output-weighted model-error criterion, which bounds (i.e. it is more conservative) the original criterion (\ref{norm11}) as well as the information-based criterion:  \begin{equation}\label{defQ}
Q[\sigma_{{y}}^2]= \int_{S_x}\frac{p_{x}(\mathbf{x})
 }{p_{{y}_0}({\mathbf{y}_0}(\mathbf{x}))}{\sigma_{ y}^2}(\mathbf{x;h})d\mathbf{x}.
\end{equation}In practice, $S_x$ is chosen as $\mathbb{R}^m$ or the support of the input pdf, $p_x$. Because of the inequality (\ref{ineq10}) we can conclude that converge of $Q[\sigma_{{y}}^2]$ also implies convergence of the metric $D_{Log^1}(\mathbf{y}_+\Vert\mathbf{y}_0)$.  However, the $Q$ criterion is much easier to compute compared with $D_{Log^1}(\mathbf{y}_+\Vert\mathbf{y}_0)$ and it can be employed even in high-dimensional input spaces. With the modified criterion the output data and their pdf is taken into account explicitly. In particular, the conditional variance (or uncertainty) of the model at each input point $\mathbf{x}$ is weighted by the probability of the input at this point, $p_{ x}(\mathbf{x})$, as well as  the inverse of the \textit{estimated} probability of the output  at the same input point, $p_{{y}_0}({\mathbf{y}_0}(\mathbf{x}))$. 

The term in the denominator comes as a result of considering the distance between the logarithms in (\ref{norm11}). If we had started with $D_{KL}(\mathbf{y}_0\Vert\mathbf{y}_+;{S_ y})$, we would have cancellation of this important term. Note that a relevant approach, based on the heuristic superposition
of the outcome and the mutual information criterion was presented in \cite{Verdinelli1992}.
However, there is no clear way how the two terms should be weighted or in
what sense the outcome can be superimposed to the information content. We  emphasize that the presented framework is not restricted to  linear regression problems and it can also be applied to Bayesian deep learning problems (a task that will not be considered in this work).  In addition, we have not made any assumption for the distribution of the input $\mathbf{x}$.

\subsubsection*{A simple demonstration}
\begin{figure}[ht]
\begin{center}
\includegraphics[width=0.88\textwidth]{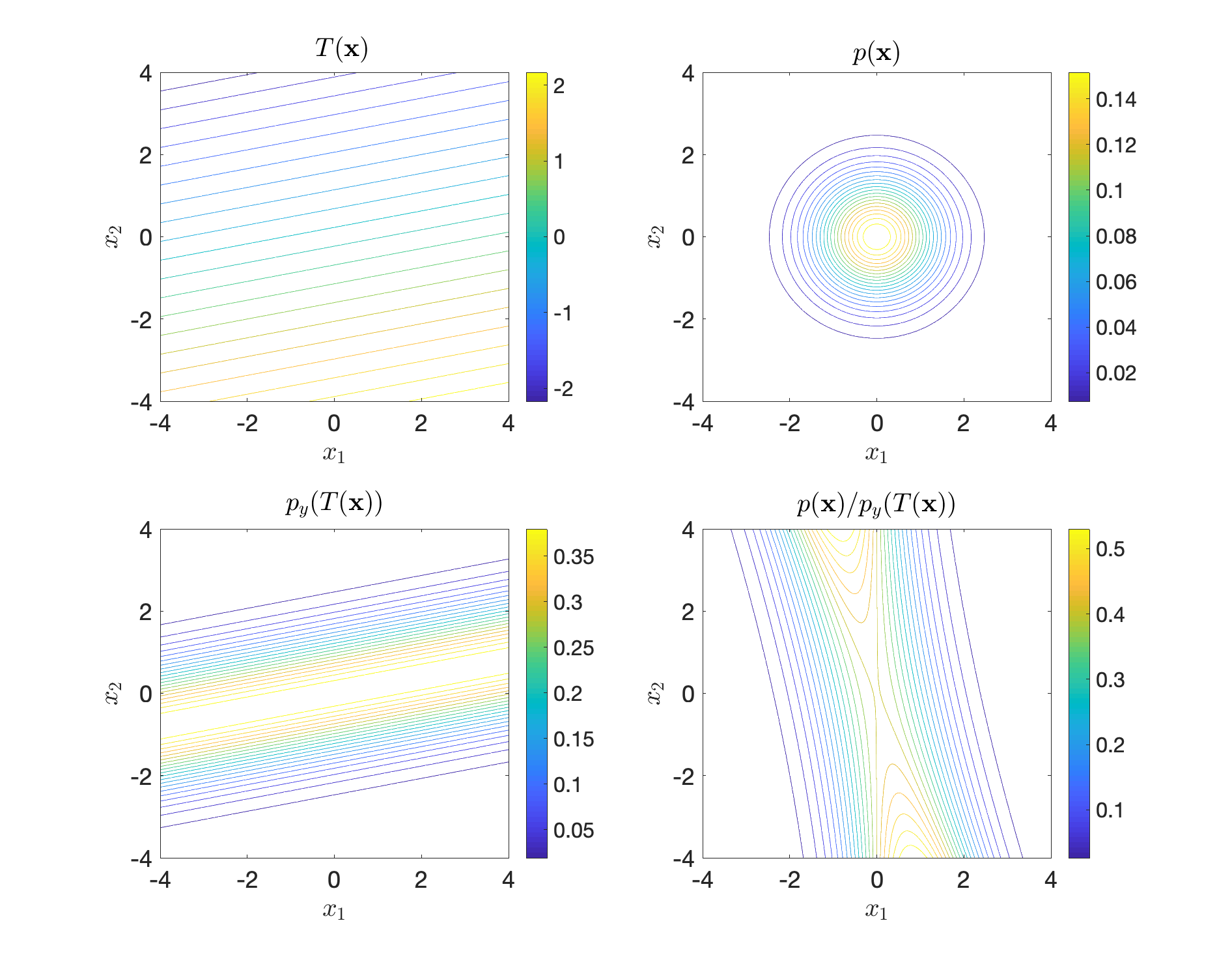}
\end{center}
\setlength{\abovecaptionskip}{-12pt}
\caption{Illustration of the criterion for sample selection. The map, $T(\mathbf{x})$,
as well as the input pdf, $p(\mathbf{x}),$ are shown in the top row, while
the conditional pdf of the output, $p_y(T(\mathbf{x}))$,  and the weight
used for sampling, $p(\mathbf{x})/p_y(T(\mathbf{x}))$, are shown in the bottom row.  }
\label{fig:framework}
\end{figure}

To illustrate the properties of the new criterion we consider the map\begin{equation*}
T(\mathbf{x})=0.1x_1-0.5x_2,  \text{ \ where \ }  \mathbf{x}\sim\mathcal{N}(0,\mathbf{I}).
\end{equation*}
Note that the $x_2$ variable is more important than $x_1$ in determining the value of the output, given that the two input variables have the same variance. It is therefore intuitive to require more accuracy for the second direction. However, the information distance or entropy based criteria take into account only the input variable statistics to select the next sample, in which case, both directions will have equal importance. This is illustrated in Figure \ref{fig:framework}, where we present contours of the exact map, $T(\mathbf{x})$, as well as, of the input pdf, $p(\mathbf{x})$. We also present the contours of the output pdf conditional on the input,  $p_{y}( T(\mathbf{x}))$ (bottom left), and the weight that is used in the criterion $Q$.  Clearly, relying on the sampling criterion that uses only information about the input will not be able to approximate the map in the most important directions. On the other hand, we observe that the weight used in the $Q$ criterion takes into account explicitly the importance of both the input variable statistics but also the information that one has estimated so far from the input-output samples. Here we used the exact map $T(\mathbf{x})$ to demonstrate the weight function but in a realistic scenario the estimated mean model $y_0(\mathbf{x})$ will be used to approximate the output pdf.

\subsection{Approximation of the criterion for symmetric output pdf}
Our efforts will now focus on the efficient approximation of the criterion $Q$.  {
To simplify the presentation we will focus on the scalar output case, $d=1$.} The first step of the approximation focuses on the denominator. This term introduces the dependence to the output data and acts as a weight to put more emphasis on regions associated with large deviations of $y(\mathbf{x})$ from its mean. We will approximate the weight, $p^{-1}_{{y}_0}({y})$, by a quadratic function that optimally represents it over the region of interest, $S_{ y}$. Therefore, for the scalar case we will have\begin{equation}
p^{-1}_{{y}_0}({y})\simeq p_1+p_2(y-\mu_y)^2,\label{eq:weights}
\end{equation} 
where $p_1, p_2$ are constants chosen so that the above expression approximates the inverse of the output pdf optimally over the region of interest. We use this expression into the definition of $Q$ (eq. (\ref{defQ})) and obtain the approximation\begin{align}\label{criterion1}
Q[\sigma_{{y}}^2] 
& \simeq p_1 \int{p(\mathbf{x})
 }{\sigma_{{y}}^2}(\mathbf{x})d\mathbf{x}+ p_2 \int{(y_{0}(\mathbf{x})-\mu_{y_0})^2p(\mathbf{x})
 }{\sigma_{{y}}^2}(\mathbf{x})d\mathbf{x}.
\end{align} 
Note that the first term does not depend on the output values  but only on the input process. It is essentially the same term that appears in the entropy based criteria. The second term however depends explicitly on the deviation of the output process from its mean and therefore on the output data. Specifically, it has large values in regions of $\mathbf{x}$ where the output process has important deviations from its mean, essentially promoting the sampling of these regions. The two constants $p_1, p_2$ provide the relative weight between the two contributions. They are computed for a Gaussian approximation of the output pdf in Appendix B. For the case where the pdf $p_y$ is expected to have important skewness, i.e. asymmetry around its mean, a linear term can be included in the expansion of $p^{-1}_{{y}_0}({y})$, so that this asymmetry is reflected in the sampling process.

\subsection{Linear regression with Gaussian input}
For the case of liner regression the first term in the criterion (\ref{criterion1}) will take the form \begin{equation}
 \int{p(\mathbf{x})
 }{\sigma_{{y}}^2}(\mathbf{x;h})d\mathbf{x}=\sigma_V^2(1+\mu_c(\mathbf{h}))=\sigma_V^2(1+\text{tr}[\mathbf{S}'^{-1}_{xx}\mathbf{R}_{xx}]),
\end{equation}
where we have considered the case of a scalar output with  $\mathbf{V}=\sigma_V^2$.  The second term of the criterion (\ref{criterion1}) will take the form\begin{align*}
 \frac{1}{\sigma_V^2}\int{(y_0(\mathbf{x})-\mu_{y_0})^2p(\mathbf{x})
 }{\sigma_{{y}}^2}(\mathbf{x;h})d\mathbf{x} & =\int{[(\mathbf{S}_{yx}\mathbf{S}_{xx}^{-1}\mathbf{(x}-\mu_{{x}})]^2
 }(1+\mathbf{x}^T\mathbf{S}'^{-1}_{xx}\mathbf{x})p(\mathbf{x})d\mathbf{x}\\
  & =c_0+\int{(\mathbf{x}-\mu_{x})^T\mathbf{S}_{xx}^{-1}\mathbf{S}^T_{yx}\mathbf{}\mathbf{S}_{yx}\mathbf{S}_{xx}^{-1}\mathbf{(x}-\mu_{x})
 }\mathbf{x}^T\mathbf{S}'^{-1}_{xx}\mathbf{x}p(\mathbf{x})d\mathbf{x}.
\end{align*}
where $c_{0}$ is a constant that does not depend on $\mathbf{h}$
\begin{equation*}
c_{0}=\int{[(\mathbf{S}_{yx}\mathbf{S}_{xx}^{-1}\mathbf{(x}-\mu_x)]^2
 }p(\mathbf{x})d\mathbf{x}=\text{tr}[\mathbf{S}_{xx}^{-1}\mathbf{S}^T_{yx}\mathbf{}\mathbf{S}_{yx}\mathbf{S}_{xx}^{-1}\mathbf{C}_{xx}].
\end{equation*}We observe that the second term depends on fourth order moments of the input process $\mathbf{x}$ but also on the output values of the samples $Y$. {This term can be computed in a closed form for the case of Gaussian input}. Specifically, \begin{align*}\int{(\mathbf{x}-\mu_x)^T\mathbf{S}_{xx}^{-1}\mathbf{S}^T_{yx}\mathbf{}\mathbf{S}_{yx}\mathbf{S}_{xx}^{-1}\mathbf{(x}-\mu_x)
 }\mathbf{x}^T\mathbf{S}'^{-1}_{xx}\mathbf{x}p(\mathbf{x})d\mathbf{x}& =\int{\mathbf{x'}^T\mathbf{S}_{xx}^{-1}\mathbf{S}^T_{yx}\mathbf{}\mathbf{S}_{yx}\mathbf{S}_{xx}^{-1}\mathbf{x'}
}\mathbf{x'}^T\mathbf{S}'^{-1}_{xx}\mathbf{x'}p(\mathbf{x'})d\mathbf{x'}\\
& +\int{\mathbf{x'}^T\mathbf{S}_{xx}^{-1}\mathbf{S}^T_{yx}\mathbf{}\mathbf{S}_{yx}\mathbf{S}_{xx}^{-1}\mathbf{x'}
}\mathbf{x'}^T\mathbf{S}'^{-1}_{xx}\mu_{x}p(\mathbf{x'})d\mathbf{x'}\\
& +\int{\mathbf{x'}^T\mathbf{S}_{xx}^{-1}\mathbf{S}^T_{yx}\mathbf{}\mathbf{S}_{yx}\mathbf{S}_{xx}^{-1}\mathbf{x'}
}\mathbf{\mu}_{x}^T\mathbf{S}'^{-1}_{xx}\mathbf{x'}p(\mathbf{x'})d\mathbf{x'}\\
& +\int{\mathbf{x'}^T\mathbf{S}_{xx}^{-1}\mathbf{S}^T_{yx}\mathbf{}\mathbf{S}_{yx}\mathbf{S}_{xx}^{-1}\mathbf{x'}
}\mathbf{\mu}_{x}^T\mathbf{S}'^{-1}_{xx}{\mu_x}p(\mathbf{x'})d\mathbf{x'},
\end{align*} 
where $\mathbf{x}'=\mathbf{x}-\mu_x$ and $p(\mathbf{x}')$ is the zero-mean translation pdf of the original one. The second and third term on the right hand side vanish as they consist of third order central moments of a Gaussian random variable. 
For the first term we employ a theorem for the covariance of quadratic forms  which gives for two symmetric matrices, $A$ and $B$    \cite{Rencher08}:
 \begin{equation*}
 \text{cov}(\mathbf{x}^TA\mathbf{x},\mathbf{x}^TB\mathbf{x})=2\text{tr}(AC_{xx}BC_{xx})+4\mu_xAC_{xx}B\mu_x.
\end{equation*}
Therefore,\begin{equation*}
\mathbb{E}[\mathbf{x}^TA\mathbf{x}\mathbf{x}^TB\mathbf{x}]=2\text{tr}(AC_{xx}BC_{xx})+4\mu_xAC_{xx}B\mu_x-\text{tr}(AC_{xx})\text{tr}(BC_{xx}).
\end{equation*}
From this equation, it follows,
 \begin{align*}
\int{\mathbf{x'}^T\mathbf{S}_{xx}^{-1}\mathbf{S}^T_{yx}\mathbf{}\mathbf{S}_{yx}\mathbf{S}_{xx}^{-1}\mathbf{x'}
}\mathbf{x'}^T\mathbf{S}'^{-1}_{xx}\mathbf{x'}p(\mathbf{x'})d\mathbf{x'}& =2\text{tr}[\mathbf{S}_{xx}^{-1}\mathbf{S}^T_{yx}\mathbf{}\mathbf{S}_{yx}\mathbf{S}_{xx}^{-1}\mathbf{C}_{xx}\mathbf{S}'^{-1}_{xx}\mathbf{C}_{xx}]\\
& -c_{0}\text{tr}[\mathbf{S}'^{-1}_{xx}\mathbf{C}_{xx}].
\end{align*}
In addition, the last term becomes
\begin{equation*}
\int{\mathbf{x'}^T\mathbf{S}_{xx}^{-1}\mathbf{S}^T_{yx}\mathbf{}\mathbf{S}_{yx}\mathbf{S}_{xx}^{-1}\mathbf{x'}
}\mu_x^T\mathbf S'^{-1}_{xx}\mu_xp(\mathbf{x'})d\mathbf{x'}=\mathbf{\mu}_{x}^T\mathbf{S}'^{-1}_{xx}{\mu_x}\text{tr}[\mathbf{S}_{xx}^{-1}\mathbf{S}^T_{yx}\mathbf{}\mathbf{S}_{yx}\mathbf{S}_{xx}^{-1}\mathbf{C}_{xx}]=c_0\mathbf{\mu}_{x}^T\mathbf{S}'^{-1}_{xx}{\mu_x}.
\end{equation*}
We collect all the computed terms and obtain
\begin{align}
\begin{split}
Q(\mathbf{h})\frac{1}{\sigma_V^2}& =p_1(1+\text{tr}[\mathbf{S}'^{-1}_{xx}\mathbf{C}_{xx}]+\mathbf{\mu}_{x}^T\mathbf{S}'^{-1}_{xx}{\mu_x})+p_{2}c_{0}(1+{\mu}_{x}^T\mathbf{S}'^{-1}_{xx}{\mu_x}-\text{tr}[\mathbf{S}'^{-1}_{xx}\mathbf{C}_{xx}])
\\ & +2p_2\text{tr}[\mathbf{S}_{xx}^{-1}\mathbf{S}^T_{yx}\mathbf{}\mathbf{S}_{yx}\mathbf{S}_{xx}^{-1}\mathbf{C}_{xx}\mathbf{S}'^{-1}_{xx}\mathbf{C}_{xx}].
\end{split}
\end{align}
This is the form of the $Q$ criterion under the assumption of Gaussian input for the case of linear regression. For the case of zero mean input it becomes
\begin{align}\label{Q_com}
Q(\mathbf{h})\frac{1}{\sigma_V^2}& =(p_1-p_{2}c_{0})\text{tr}[\mathbf{S}'^{-1}_{xx}\mathbf{C}_{xx}] +2p_2\text{tr}[\mathbf{S}_{xx}^{-1}\mathbf{S}^T_{yx}\mathbf{}\mathbf{S}_{yx}\mathbf{S}_{xx}^{-1}\mathbf{C}_{xx}\mathbf{S}'^{-1}_{xx}\mathbf{C}_{xx}]+\text{const.}
\end{align}
The coefficients $p_1, p_2$ are determined using the output pdf of the estimated model through the samples $D$ (eq. (\ref{eq:weights})), i.e. the pdf of $y_0(\mathbf{x})$. Note that the exact form of the output pdf, used in the criterion, is not important at this stage as it only defines the weights of the criterion $Q(\mathbf{h})$. For a Gaussian approximation of the output process the coefficients are given in Appendix B. 
\subsection{Nonlinear regression with Gaussian input}

For the case of regression with non-linear basis the first term in the criterion (\ref{criterion1})
will take the form \begin{equation}
 \int p(\mathbf{z})
 {\sigma_{{y}}^2}(\mathbf{z;h})d\mathbf{z}=\sigma_V^2(1+\mu_c(\mathbf{h}))=\sigma_V^2(1+\text{tr}[\mathbf{S}'^{-1}_{\phi\phi}\mathbf{C}_{\phi\phi}]+\mathbf{\mu}_{\phi}^T\mathbf{S}'^{-1}_{\phi\phi}{\mu_\phi}),
\end{equation}
where we have considered the case of a scalar output with  $\mathbf{V}=\sigma_V^2$.
 The second term of the criterion (\ref{criterion1}) will take the form\begin{align*}
 \frac{1}{\sigma_V^2}\int{(y_0(\mathbf{z})-\mu_{y_0})^2p(\mathbf{z})
 }{\sigma_{{y}}^2}(\mathbf{z;h})d\mathbf{x} & =\int{[(\mathbf{S}_{y\phi}\mathbf{S}_{\phi\phi}^{-1}\mathbf{(\phi(z)}-\mu_{{\phi}})]^2
 }(1+\phi(\mathbf{z})^T\mathbf{S}'^{-1}_{\phi\phi}\phi(\mathbf{z}))p(\mathbf{z})d\mathbf{z}\\
  & =\int{[(\mathbf{S}_{y\phi}\mathbf{S}_{\phi\phi}^{-1}\mathbf{(\phi}-\mu_{{\phi}})]^2
 }(1+\phi^T\mathbf{S}'^{-1}_{\phi\phi}\phi)p(\phi)d\phi,
\end{align*}
where we expressed the integral using the pdf for the basis elements $\phi$. In this way the integral is now expressed exactly as in the linear regression case. To obtain a closed approximation we approximate the pdf for $\phi$ through its second-order statistics (i.e. approximate $p(\phi)$ with a Gaussian pdf). The analysis shown for the linear case with Gaussian input is then valid leading to the following expression for the $Q$ criterion:
\begin{align}
\begin{split}
Q(\mathbf{h})\frac{1}{\sigma_V^2}& =p_1(1+\text{tr}[\mathbf{S}'^{-1}_{\phi\phi}\mathbf{C}_{\phi\phi}]+\mathbf{\mu}_{\phi}^T\mathbf{S}'^{-1}_{\phi\phi}{\mu_\phi})+p_{2}c_{0}(1+{\mu}_{\phi}^T\mathbf{S}'^{-1}_{\phi\phi}{\mu_\phi}-\text{tr}[\mathbf{S}'^{-1}_{\phi\phi}\mathbf{C}_{\phi\phi}])
\\ & +2p_2\text{tr}[\mathbf{S}_{\phi\phi}^{-1}\mathbf{S}^T_{y\phi}\mathbf{}\mathbf{S}_{y\phi}\mathbf{S}_{\phi\phi}^{-1}\mathbf{C}_{\phi\phi}\mathbf{S}'^{-1}_{\phi\phi}\mathbf{C}_{\phi\phi}].
\label{eq:q_NL}
\end{split}
\end{align}
So, for a given basis $\phi(\mathbf{z})$ one needs first to obtain the mean vector  $\mu_\phi$ and covariance $C_{\phi\phi}$ using the expressions (\ref{mu_cor_phi}) and then follow the same steps as in the linear case. The expression for the gradient of the $Q$ criterion, under general choice of $\phi(\mathbf{z})$ is given in Appendix A. 
\section{Examples}
\subsection{ Linear map with a 2d input space}\label{linear2d}

\begin{figure}[ht]
\begin{center}
\includegraphics[width=0.45\textwidth]{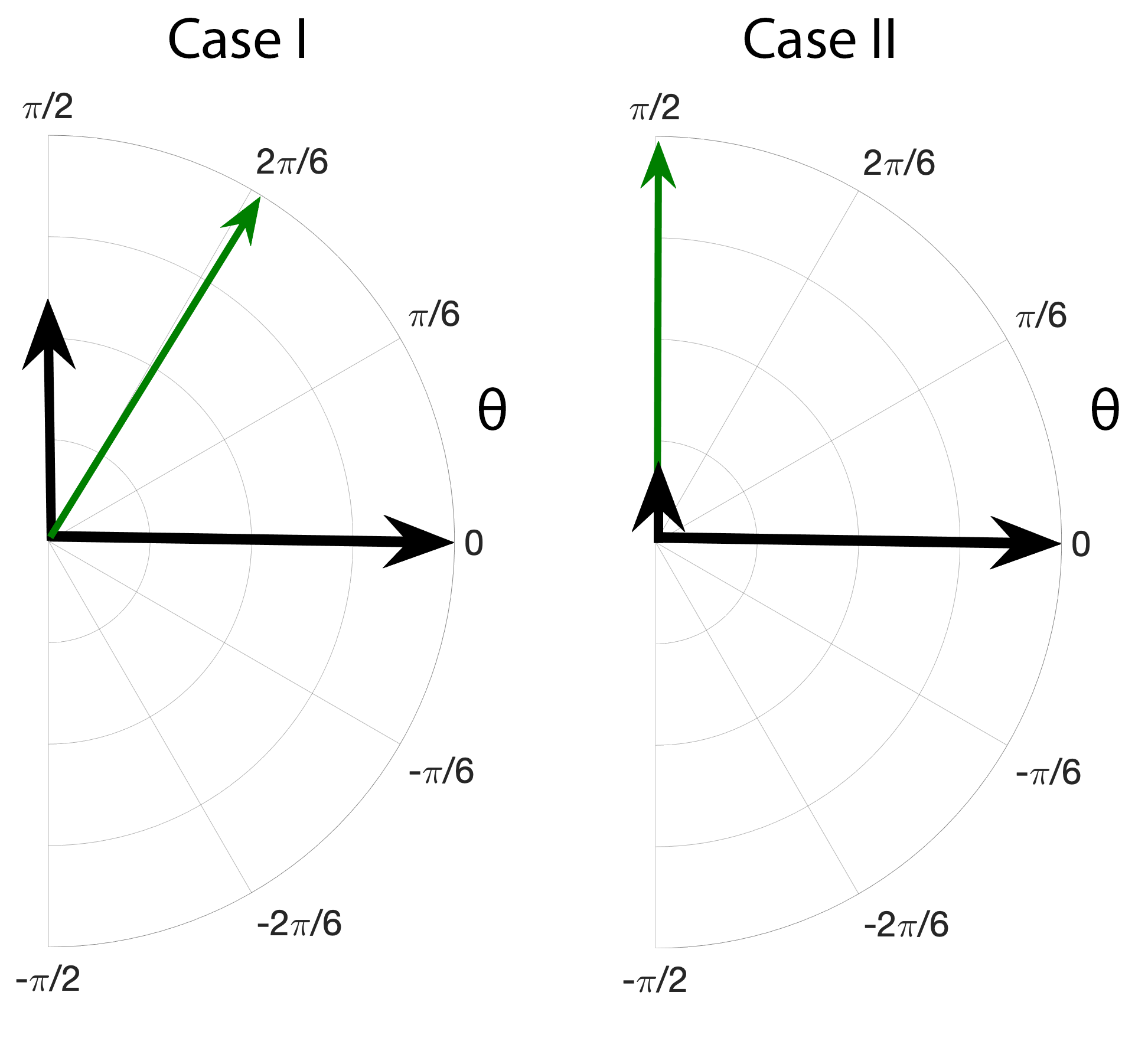}
\end{center}
\setlength{\abovecaptionskip}{-12pt}
\caption{Black arrows indicate direction and magnitude of the principal directions (and corresponding eigenvalues) of the input covariance $\mathbf{C}_{xx}$ for each case of parameters. The green arrow indicates the gradient of the map $T(\mathbf{x})$. }
\label{fig_2d_cases}
\end{figure}

To demonstrate the properties of the new criterion we first consider the two-dimensional  problem\begin{equation}
T(\mathbf{x})=\hat a_{1}x_1+\hat a_{2}x_2 + \epsilon,  \text{ \ where \ }  \mathbf{x}\sim\mathcal{N}(0,\mathbf{\begin{bmatrix}\sigma_1^2 & 0 \\
0 & \sigma_2^2 \\
\end{bmatrix}}) \text{ \ and \ } \sigma_V^2=0.05.
\end{equation}
We consider two cases of parameters\begin{itemize}
\item 
Case I\ :  $\hat a_1=0.8, \hat a_2=1.3$, and $\sigma_1^2=1.4, \sigma_2^2=0.6.$
\item 
Case II: $\hat a_1=0.01, \hat a_2=2.0$, and $\sigma_1^2=2.0, \sigma_2^2=0.2.$
\end{itemize}  
The two cases are presented in Figure
\ref{fig_2d_cases} in polar coordinates. The black arrows indicate the principal
directions of the input covariance, scaled according to the eigenvalues of
the covariance matrix, while the green arrow indicates the direction of the
gradient of the map $T(\mathbf{x})$. While for the first case the contributions of both input variables to the output are comparable and thus sampling is important for both of them, for Case II the contribution of the first input variable is negligible. However, this input variable is the one with the highest uncertainty. 

  For each case we assess four adaptive sampling strategies according to the criteria:  \begin{enumerate}
\item 
The directly computed mutual information,  $\mathcal{I}(\mathbf{x,y}|D')$ is maximized in $\mathbb{S}^1$,
\item The second-order statistical approximation of the mutual information,  $\mathcal{I}_G(\mathbf{x,y}|D')$ is maximized in $\mathbb{S}^1$,
\item The uncertainty parameter,  $\mu_c(\mathbf{h})$ is minimized in $\mathbb{S}^1$, \item The output-weighted model error criterion
 $Q(\mathbf{h})$ is minimized in $\mathbb{S}^1$.  
 \end{enumerate}
 For the $Q$ criterion we choose $p_1=0$ and $ p_2=1$ to emphasize the role of the second, new term that takes into account the output samples. This case of parameters corresponds to the case where we optimally approximate $p_{y_0}^{-1}$ over the full real axis, i.e. $\beta \rightarrow\infty$ using the notation of Appendix B. We denote this criterion as $Q_{\infty}$. 
  We also  compare with a Monte-Carlo approach where samples are randomly
generated from the input distribution of $\mathbf{x}$ and then normalized
so they belong in $\mathbb{S}^1$. 

For the adaptive strategies based on $\mu_c$ and $Q_{\infty}$ we use the analytical
expressions (\ref{eq_mc}) and (\ref{Q_com}), respectively together with their gradient computed in Appendix A. This allowed us to utilize gradient based optimization methods. For the adaptive strategies based on the mutual information, $\mathcal{I}(\mathbf{x,y}|D,\mathbf{h})$
and its second-order approximation,  $\mathcal{I}_G(\mathbf{x,y}|D,\mathbf{h})$, we used a random sampling approach and equations (\ref{info_full}) and (\ref{mutualc}), respectively. Specifically, we generated $10^5$ samples from the input distribution $\mathbf{x}$ and utilized the exact
expression:\begin{equation*}
\mathbb{E}^{x}[\mathcal{E}_{y|x}(\mathbf{x;h})]=\mathbb{E}^{x}[\log(1+\mathbf{x}^T\mathbf{S}_{xx}'^{-1}\mathbf{x})],
 \end{equation*} which was  numerically computed as an ensemble average.
For the computation of $\mathcal{I}$ we also generated $10^5$ realizations of the vector $\mathbf{a}=(a_1,a_2)$,
which according to the linear regression method follows the normal distribution
(\ref{coef_dis}):\begin{equation*}
p(\mathbf{a}|D',\sigma_V^2) \sim\mathcal{N}(\mathbf{S}_{yx}\mathbf{S}_{xx}^{-1},\sigma_V^2\mathbf{S}_{xx}'^{-1}).
\end{equation*}
Next, we compute a pdf approximation for $y$ using the generated samples and the kernel smoothing functions method \cite{hill85abc},
and we approximated the entropy of the resulted distribution by direct numerical
integration. 
Note that this additional step, required for  $\mathcal{I}$, has a vast  computational cost. Most importantly, because of the absence of an analytical expression for the gradient of $\mathcal{I}$, its application to high dimensional inputs is impossible. For this example the next sample vector was parametrized as $\mathbf{h}=[\cos(\theta),\sin(\theta)]$ and the criterion was optimized by direct selection of the maximum value over a one dimensional grid for $\theta \in [-\frac{\pi}{2},\frac{\pi}{2}]$.   

\begin{figure}[ht]
\begin{center}
\includegraphics[width=0.95\textwidth]{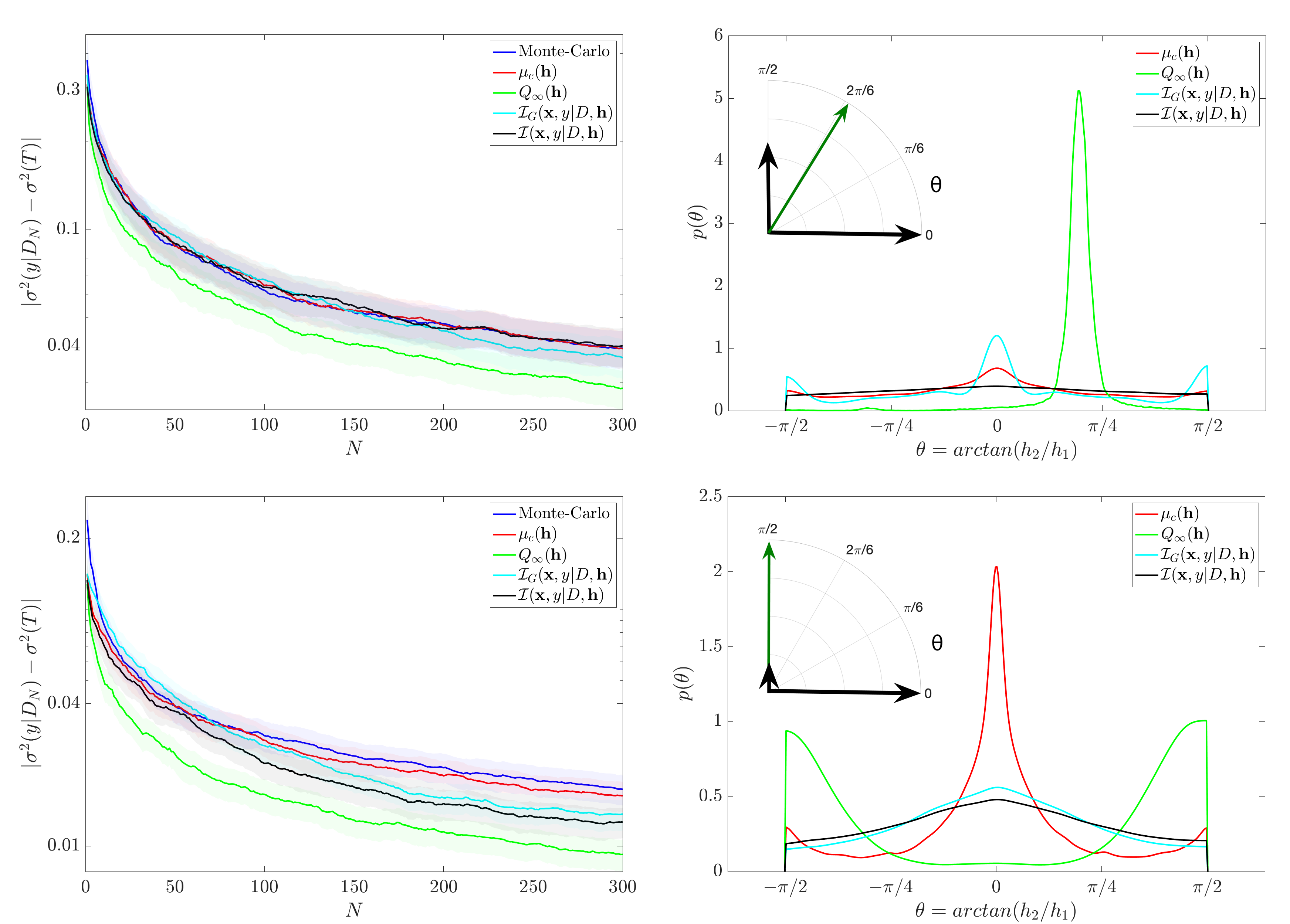}
\end{center}
\caption{Comparison of the four adaptive strategies based on different criteria
and the Monte-Carlo method. On the left plots the average error of the output
variance is shown with respect to the number of samples used over 400 experiments
for each criterion. The shaded regions indicate $0.2\sigma$ based on the
400 numerical experiments.  The pdf of these samples is shown for each adaptive
strategy in the plots on the   right. The black vectors indicate the eigenvectors
of the input covariance $\mathbf{C}_{xx}$ and the green vector denotes the
gradient of the exact map: $(\hat a_1,\hat a_2)$.}
\label{fig:2d}
\end{figure}

 All four adaptive strategies are initiated with $4$ random samples drawn from the $\mathbf{x}$ distribution. For each case we present the average error curve over $400$ experiments, i.e. experiments with different sets of initial samples (Figure \ref{fig:2d}: left panels). The standard deviation for each error curve is also presented in the shaded region. For the four adaptive sampling strategies we also present the pdf of the orientation of the samples $\mathbf{h}=(h_1,h_2)$, i.e. $\theta=\arctan(h_2/h_1)$. 

In both cases of parameters we observe that the strategy based on the $Q_{\infty}$ criterion outperforms the other three adaptive strategies. This difference in performance is even more pronounced for Case II, where one of the input variables has negligible contribution but large uncertainty. An interesting observation is that the Monte-Carlo strategy, performs as good as the $\mu_c$ criterion and the mutual information criteria, $\mathcal{I}$ and $\mathcal{I}_G$. This is not a surprise given that   $\mathcal{I}_G$ depends primarily on $\mu_c$ and the latter is designed to give more emphasis on input directions with large uncertainty, without taking into account their expected contributions to the output, similarly with Monte-Carlo. The same conclusions
hold even for the full mutual information criterion, an indication that although
$\mathcal{I}$ partially incorporates the output samples, it does not do it
in a useful way.    

Similar observations can be made if we examine the pdf for the input samples obtained from the four adaptive strategies. We can see that for each case of parameters the strategies based on $\mu_c$, $\mathcal{I}$, and $\mathcal{I}_G$  behave very similarly and tend to place input samples in the direction of larger uncertainty.  On the other hand, the $Q_{\infty}$ based strategy is placing more samples in directions that compromise between  large expected impact to the output variable but also with important uncertainty. {We note that all the cases presented here  correspond to a known output variance, $\sigma_V^2$. Results for this example corresponding to unknown
variance $\sigma_V^2$ are presented in Appendix D.}
\subsection{A high-dimensional linear problem}
 The next problem to demonstrate the optimal sampling approach is a 20 dimensional linear function. Note that for this case optimization of the mutual information is an impossible task given the fact that the full expression for the mutual information is hard to optimize in the absence of expressions for its gradient. 

\begin{figure}[h]
\begin{center}
\includegraphics[width=0.88\textwidth]{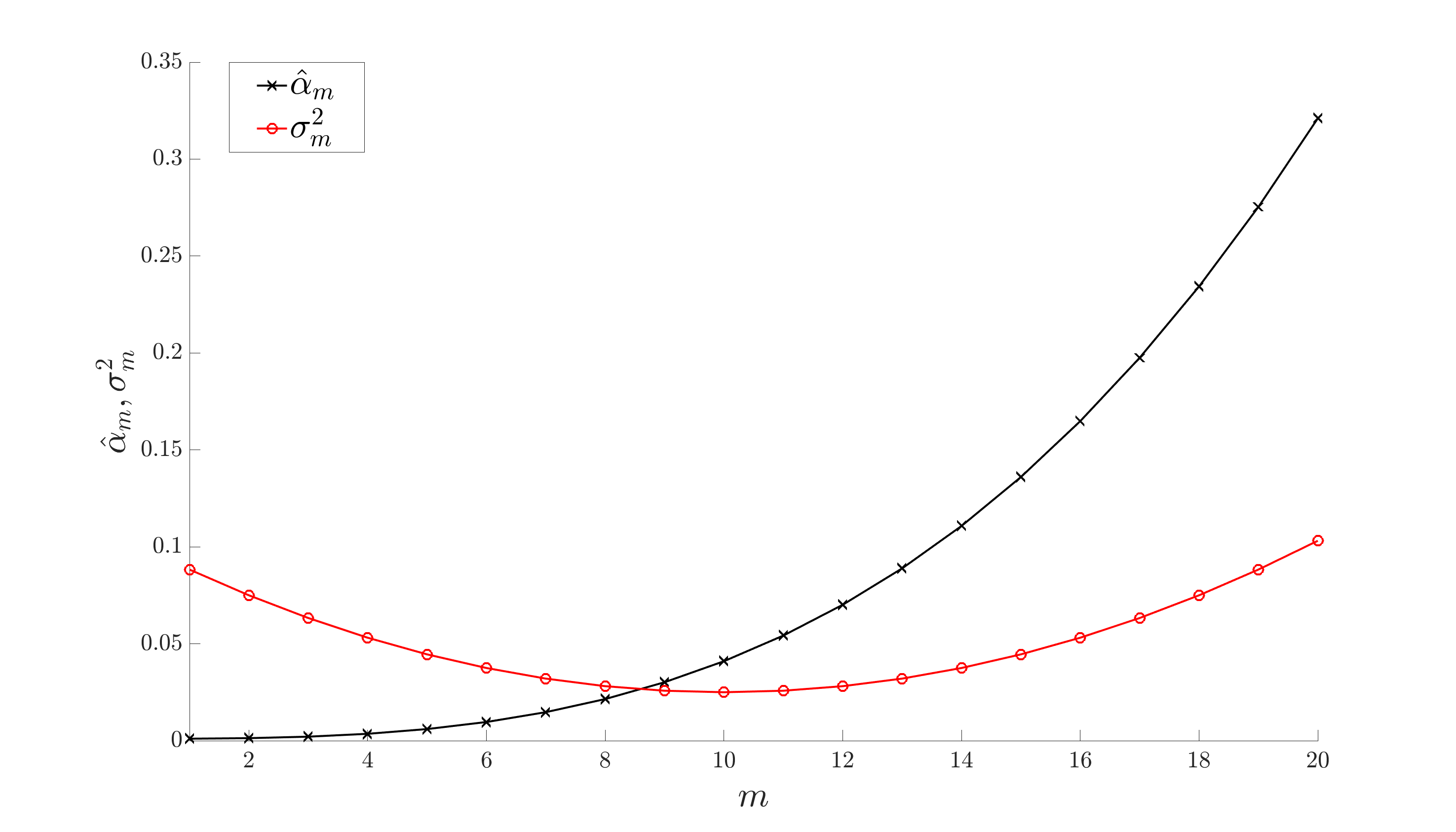}
\end{center}
\setlength{\abovecaptionskip}{0pt}
\caption{Coefficients, $\hat \alpha_m$, of the 
map $T(\mathbf{x})$ (black curve) plotted together with the variance of each input direction $\sigma_m^2$ (red curve) for the high dimensional problem (\ref{eq:highd}).}
\label{fig_highD}
\end{figure}

Specifically,  we consider the  system
\begin{equation}
T(\mathbf{x})=\sum_{m=1}^{20}\hat a_{m}x_m + \epsilon,  \text{ \ where
\ }  {x_m}\sim\mathcal{N}(0,\sigma_m^2 ), \ \  m=1,...,20,\label{eq:highd}
\end{equation}
where the coefficients and input variances are chosen as\begin{align*}
\hat a_{m}& =\left(1+40\left( \frac{m}{10} \right)^3 \right) 10^{-3}, \ m=1,...,20,\\
\sigma_m^2& =\left(\frac{1}{4}+\frac{1}{128}\left( m-10 \right)^3 \right) 10^{-1}, \ m=1,...,20.
\end{align*}
This system represents a typical high dimensional case, where we have some very influential degrees of freedom and some that have negligible impact to the output variable. The energy of these coefficients is typically not related to their influence to the output variable. In Figure \ref{fig_highD} we present the coefficients and input variances.  

For the observation noise we consider two cases:\begin{itemize}
\item 
Case I: $\sigma_{\epsilon}^2=0.05$ (accurate observations)
\item
Case II: $\sigma_{\epsilon}^2=0.5$ (noisy observations)
\end{itemize}  
Given that $\sum_{m=1}^{20}\hat a_{m}^2 \sigma_m^2=0.0272$ the first case corresponds to relatively accurate observations while the second is a highly noisy observations case. We expect the adaptive sampling approach to be more valuable for the second case, given that for the first case we need very few samples anyway. 

We apply the adaptive criteria after we have obtained one sample per input direction, to guarantee that the matrix $\mathbf{S}_{xx}$ is invertible. Then we run each numerical experiment $L=400$ times to make sure that the randomness due to the observation noise does not favor any method.

\begin{figure}[h]
\begin{center}
\includegraphics[width=0.98\textwidth]{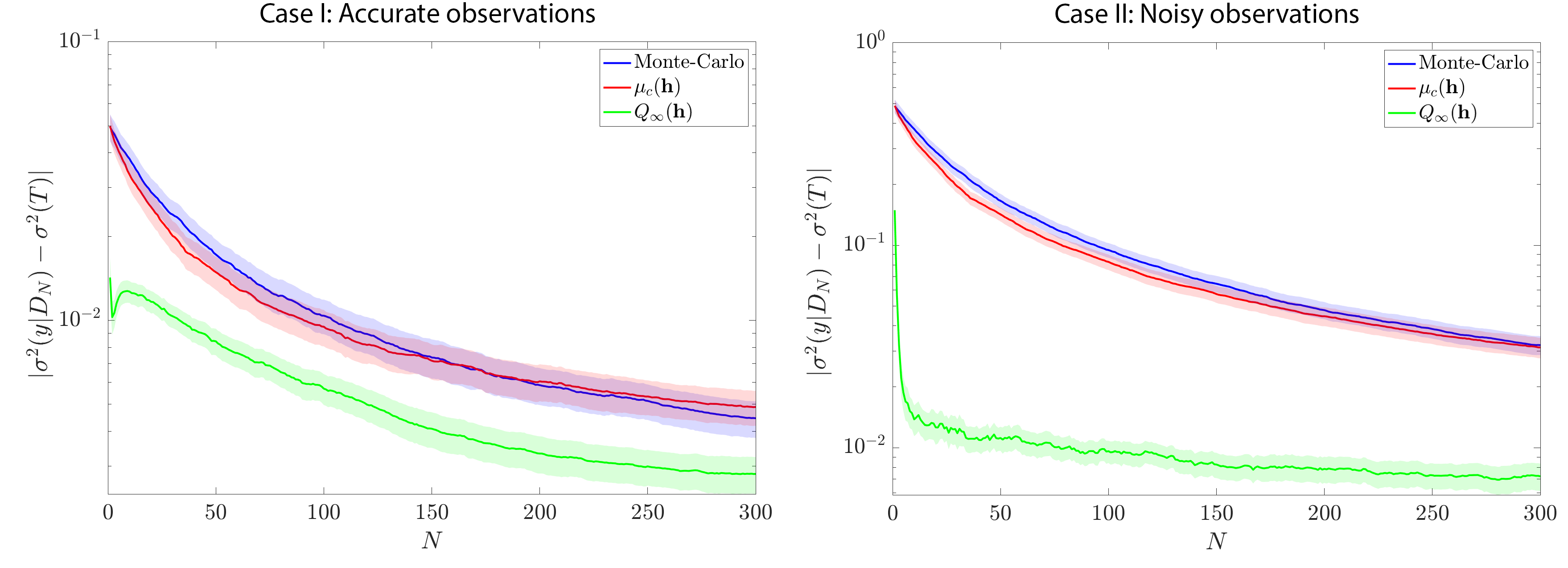}
\end{center}
\setlength{\abovecaptionskip}{-0pt}
\caption{Performance of the two adaptive approaches based on $\mu_c$ and
$Q_{\infty}$ for the high dimensional problem (\ref{eq:highd}). The left plot corresponds to observation noise, $\sigma_{\epsilon}^2=0.05$ (accurate
observations) and the right to $\sigma_{\epsilon}^2=0.5$ (noisy observations).}
\label{fig_highD_curves}
\end{figure}

 In Figure \ref{fig_highD_curves} we present the performance of the sampling approach based on $\mu_c$ and $Q_{\infty}$, as well as a direct Monte-Carlo approach. For the first case (accurate observations), shown in the left plot, we note a clear advantage of the $Q_{\infty}$ sampling approach that takes into account the output samples. This advantage is more pronounced in the second case of noisy observations, where the approach using the  $Q_{\infty}$ criterion obtains an order of magnitude higher accuracy from the very first samples. Note that the $\mu_c$ sampling strategy  is comparable with the Monte-Carlo approach, since it does not take into account the output samples.

\begin{figure}[t]
\begin{center}
\includegraphics[width=0.88\textwidth]{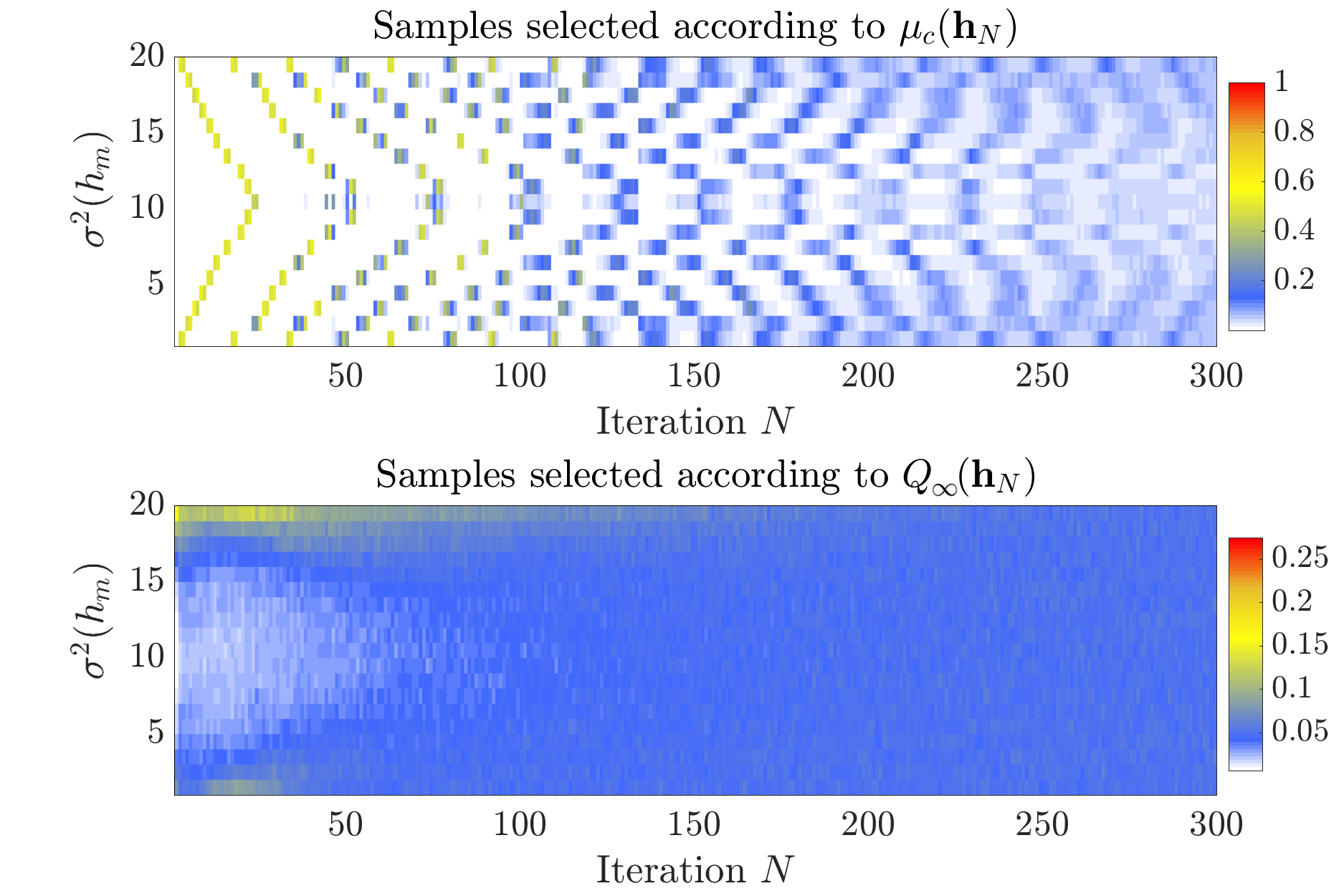}
\end{center}
\setlength{\abovecaptionskip}{-0pt}
\caption{Energy of the different components of $\mathbf{h}$ with respect
to the number of iteration $N$ for Case I of the high dimensional problem.}
\label{fig_highD_samples}
\end{figure}

The same conclusions can be obtained if we observe the variance of the $\mathbf{h}_N$ components, over different runs of the numerical experiments, $l=1,...,L$  (here $L=400$), i.e. over different realizations of the observation noise:\begin{equation}
\sigma^2(h_m)=\frac{1}{L}\sum_{l=1}^L\left( h_{N,m,l} - \bar h_{N,m}\right)^2, \ \ \text{where} \ \ \bar  h_{N,m}=\frac{1}{L}\sum_{l=1}^L h_{N,m,l}, \ \  m=1,...,20.
\end{equation}
Results are shown in Figure $\ref{fig_highD_samples}$. Sampling according to $\mu_c$ results in a distribution that is following the shape of the variance $\sigma_m$. Specifically, the scheme iteratively changes input directions based on their variance, starting from the most energetic ($m=1$ and $m=20$) and moving towards the less energetic ones ($m=10$). Then the loop begins again, until all the input directions are equally well sampled, after which point the sampling is random. 

Sampling according to $Q_{\infty}$, on the other hand, is performed in one loop starting from the most energetic directions, but giving more emphasis in the input directions close to $m=20$ that have both high energy and large contribution to the output $y$. This `asymmetry' in the sampling results in significantly faster convergence compared with the Monte-Carlo method or the $\mu_c$ criterion. 

The effect of the domain selection $S_y$ is discussed in detail in Appendix C, where a parametric study is also shown. Specifically, if we utilize a finite number of standard deviations to optimally approximate $p_{y_0}^{-1}$ (eq. (\ref{eq:weights})), by employing $Q_{\beta}$ (with $\beta$ finite), the term $\mu_c$ in the criterion improves the behavior of sampling for large $N$ (Figure \ref{fig_highD_curves_more}).  
 
\subsection{A 2d nonlinear problem with nonlinear basis functions}
\begin{figure}[t]
\begin{center}
\includegraphics[width=0.84\textwidth]{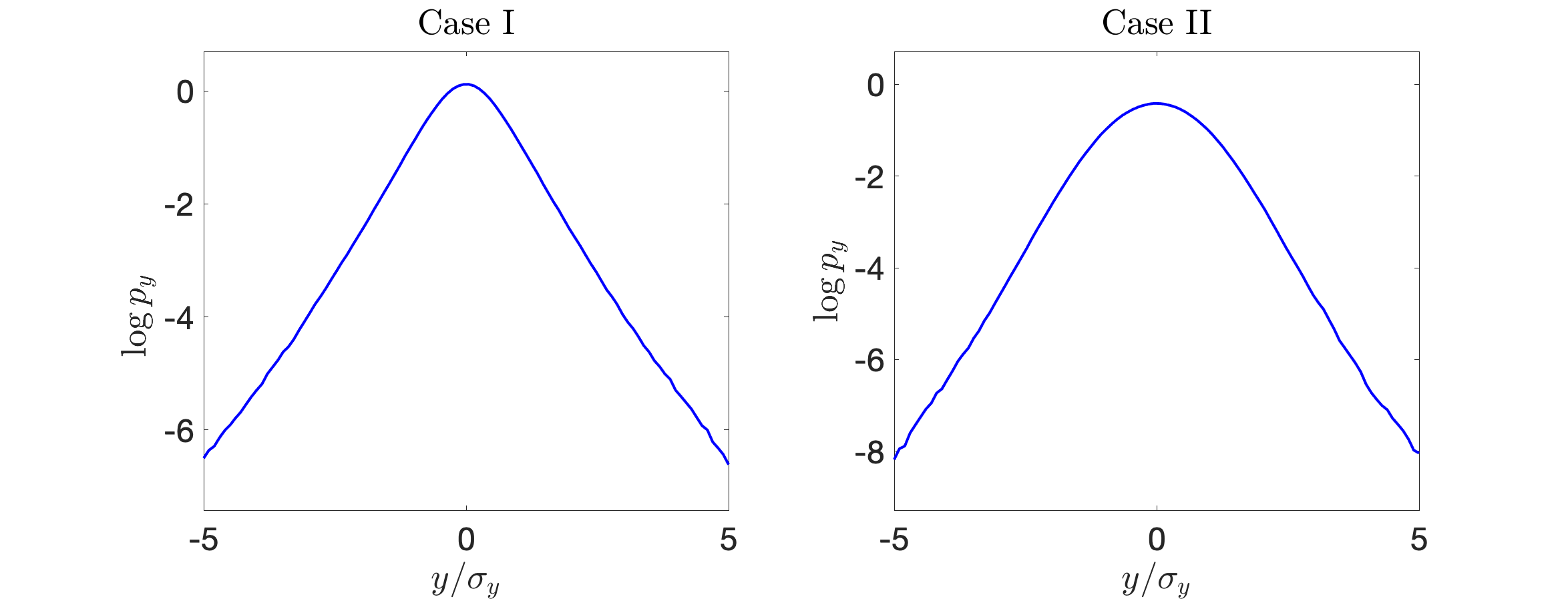}
\end{center}
\setlength{\abovecaptionskip}{-0pt}
\caption{Exact pdf for the two cases of the nonlinear map using MC with $10^5$
samples.}
\label{figNL2D_pdfs}
\end{figure}

The next application involves a nonlinear map with a 2D input space. Specifically, we consider the
two-dimensional nonlinear  problem\begin{equation}
T(\mathbf{z})=\hat a_{1}z_1+\hat a_{2}z_2 +\hat a_{3}z_1^3+\hat a_{4}z_2^3 + \epsilon,  \text{ \ where
\ }  \mathbf{x}\sim\mathcal{N}(0,\mathbf{\begin{bmatrix}\sigma_1^2 & 0 \\
0 & \sigma_2^2 \\
\end{bmatrix}}) \text{ \ and \ } \sigma_V^2=10^{-4}.
\end{equation}
We consider two cases of parameters\begin{itemize}
\item 
Case I :  $\hat a_1=10^{-2}, \hat a_2=5, \hat a_3=0, \hat a_4=10^{2}$, and $\sigma_1^2=2.10^{-1}, \sigma_2^2=5.10^{-3}.$
\item 
Case II:  $\hat a_1=10, \hat a_2=5, \hat a_3=0, \hat a_4=10^{2}$, and $\sigma_1^2=2.10^{-3},
\sigma_2^2=5.10^{-3}.$
\end{itemize}  
In the first case the output has very weak dependence on the first variable although the latter has very large variance. Moreover, the second variable has significantly smaller variance but plays the dominant role for the output. On the other hand, for the second case, both input variables play an important role and their variance is also comparable. The exact pdf computed with an expensive Monte-Carlo simulation is shown in Figure \ref{figNL2D_pdfs}. Both distributions are characterized by heavy tails. 

We setup a non-linear Bayesian regression scheme with the following odd basis functions: \begin{equation}
\phi(\mathbf{z})=z_1^iz_2^j, \ \ \ (i,j) \in \left\{ (0,1),(1,0),(1,1),(0,3),(3,0) \right\}.
\end{equation} 
This set of basis functions contains all the odd monomials with order less or equal to 3. We observe that for the nonlinear case although the input space is two-dimensional, the regression is performed in a five-dimensional space. To avoid an ill-conditioned matrix $\mathbf{S}_{\phi\phi}$ we assume a prior with covariance $\mathbf{K}=\alpha \mathbf{I}$ (see eq. (\ref{eq:prior})). For the cases considered we set $\alpha=10^{-1}$. For each case we first choose randomly (using the distribution of $\mathbf{z}$) two samples. Then we use the criteria based on $\mu_c$ (equation (\ref{eq:mu_NL})) and $Q_{0.01}$ (equation (\ref{eq:q_NL})) to optimize 100 samples. For each step we employed a gradient-based optimization using the expressions presented in Appendix A and we restricted the samples to in the disk: $|\mathbf{z}| \le 2$. For each criterion we performed 200 optimization cycles, i.e. we computed for each criterion the full sequence of 100 samples, 200 times and we computed the statistics of the error (mean and standard deviation), so that the results are not sensitive on the randomness due to observational noise or the initial samples.   

\begin{figure}[t]
\begin{center}
\includegraphics[width=0.88\textwidth]{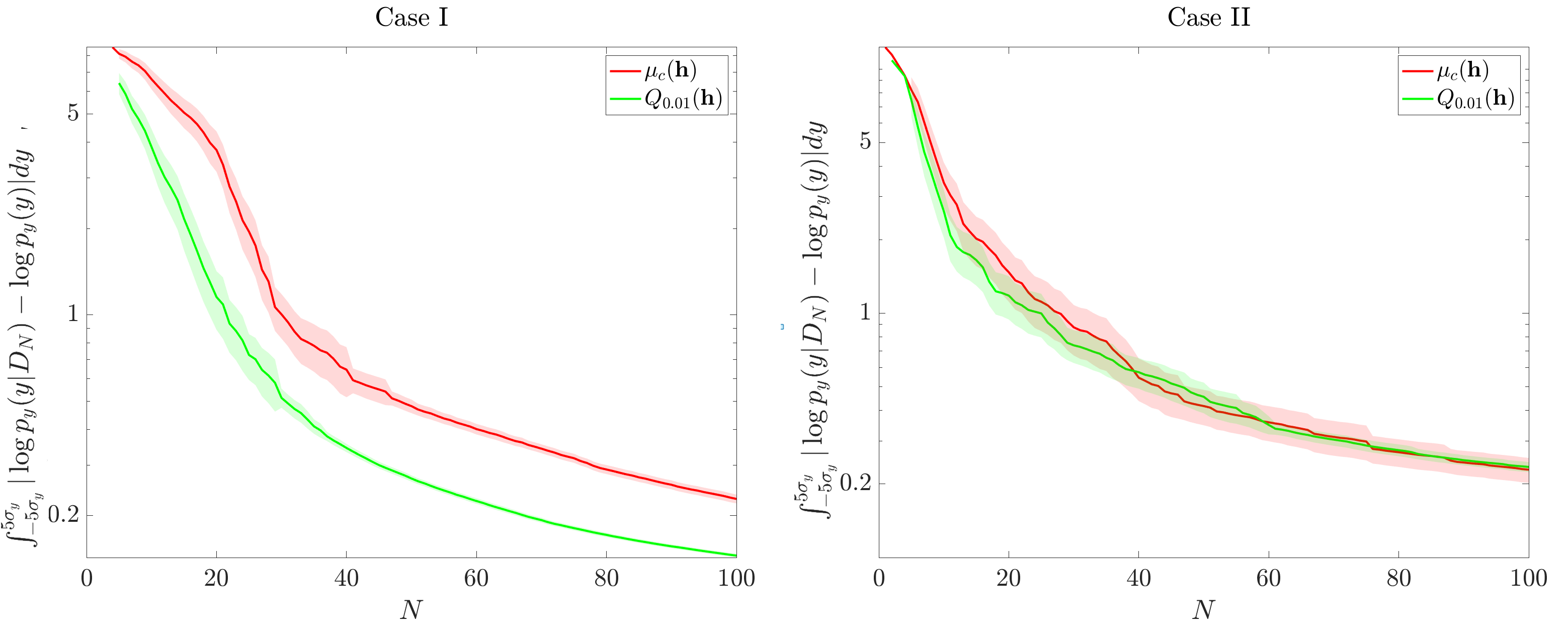}
\end{center}
\setlength{\abovecaptionskip}{-0pt}
\caption{Performance of the two adaptive approaches based on $\mu_c$ and
$Q_{\infty}$ for the nonlinear, two-dimensional problem. In the second case both input directions have important role to the output and the two methods are comparable, as expected.}
\label{figNL2D_curves}
\end{figure}

The convergence analysis for each criterion is presented in Figure \ref{figNL2D_curves}. The left plot shows the convergence of the two methods for the parameters of the first numerical experiment (Case I). Each curve is the mean error computed from the 200 optimization cycles, while the shaded area indicates the spread across different runs. Note that for Case I  there is only input variable ($z_2$)\ that plays a dominant role on the output, while the other input variable has negligible effect but important variance. In agreement with the results of the linear problems, the samples based on the $Q_{0.01}$ criterion, achieve better performance as they rapidly align with the direction that has the most important influence on the output. 

\begin{figure}[t]
\begin{center}
\includegraphics[width=0.98\textwidth]{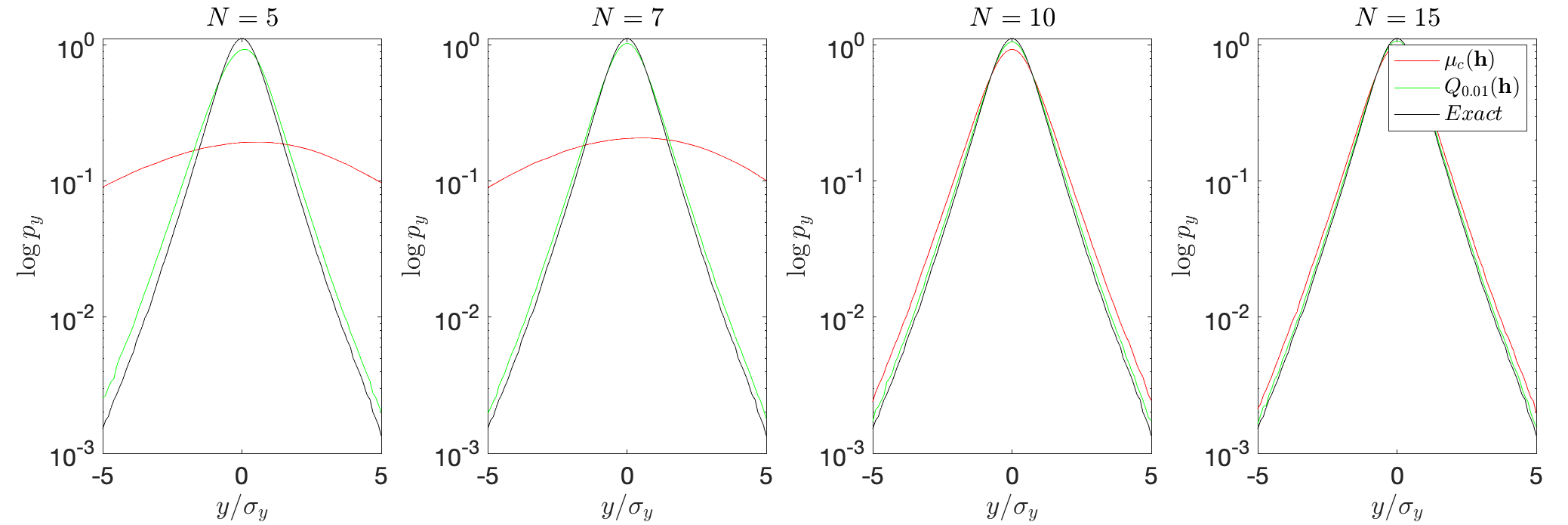}
\end{center}
\setlength{\abovecaptionskip}{-0pt}
\caption{Performance of the two adaptive approaches based on $\mu_c$ and
$Q_{\infty}$ for the nonlinear, two-dimensional problem and Case  I parameters. The resulted pdfs for samples selected according to the two criteria are compared with the exact pdf.}
\label{fig_convNL2D_curves}
\end{figure}

This is not the case for the samples based on the $\mu_c$ criterion that align primarily with the directions of importance variance, resulting in a slower convergence. For the Case II parameters both input directions have comparable variance and comparable effect to the output. In this case, as expected, the two criteria have comparable performance. This is clearly demonstrated by the right plot in Figure \ref{figNL2D_curves}.
Finally, in Figure \ref{fig_convNL2D_curves} we demonstrate the convergence of the pdfs for the first case of parameters. 

\section{Conclusions}
We have analyzed fundamental limitations of popular criteria for samples selection, employed in the optimal experimental design community. These criteria are based on maximization of entropy-based quantities, typically having the form of mutual information between input and output variables. Specifically, we have shown that beyond the large computational cost associated with these criteria that restricts their applicability to very low-dimensional problems, there is weak dependence of the induced sampling process to the output values of the existing samples { when the variance of the output noise is assumed to be known. Even for the case of unknown variance though the dependence on the output values is not controllable and can become very weak.} In this way, directions of the parameter space that contribute the most to the output may not be emphasized.

Motivated by these limitations, we have presented a new criterion for optimally selecting training samples that significantly accelerates the convergence of Bayesian regression schemes with respect to the state of the art. The criterion explicitly takes into account the fact that different input parameters have different impact to the output of interest, with some of them being much more influential than others. In this way, it places more samples towards the influential parameters, which are also characterized by important uncertainty. In addition, the introduced criterion is more practical to compute, compared with mutual information criteria, as its gradient can be analytically derived, allowing for the employment of gradient optimization methods.  Therefore, the new method allows for the optimization of samples, even for a large number of input parameters, paving the way for optimal experimental design and active learning in high-dimensions.

Future work will focus on the formulation of the presented framework on the  training of deep neural networks. The presented approach is expected to have important impact in application areas such as optimal experimental design for systems where very few experiments are available (e.g. biology), adaptive sampling in complex environments with multiple objectives, uncertainty quantification and extreme event statistics in challenging problems such as fatigue-crack, coastal flooding, critical network events, and others.  
\subsubsection*{Acknowledgments}
The author would like to thank Prof. Munther Dahleh, Prof. Sanjoy Mitter and Dr Mustafa Mohamad for several stimulating discussions. This work was initiated during a sabbatical visit of the author at ETH, hosted by Prof. George Haller, which is greatfully acknowledged. It was later supported by a Doherty Career Development Chair, a Mathworks Faculty Research Innovation Fellowship and the ARO\ MURI Grant W911NF-17-1-0306. The detailed comments of the referees led to several improvements and are also highly appreciated.
\bibliographystyle{abbrv}
\bibliography{library}

\begin{thebibliography}{10}

\bibitem{Uhler19}
R.~Agrawal, C.~Squires, K.~Yang, K.~Shanmugam, and C.~Uhler.
\newblock {ABC-Strategy: Budgeted experimental design for targeted causal
  structure discovery}.
\newblock In {\em Proceedings of Machine Learning Research 89 (AISTATS 2019)},
  page (to appear), 2019.

\bibitem{Blonigan19}
P.~J. Blonigan, M.~Farazmand, and T.~P. Sapsis.
\newblock {Are extreme dissipation events predictable in turbulent fluid
  flows?}
\newblock {\em Physical Review Fluids}, 4(4):044606, apr 2019.

\bibitem{brunton2019a}
S.~Brunton, B.~Noack, and P.~Koumoutsakos.
\newblock {Machine Learning for Fluid Mechanics}.
\newblock {\em Ann. Rev. Fluid Mech.}, In Press, 2019.

\bibitem{Chaloner95}
K.~Chaloner and I.~Verdinelli.
\newblock {Bayesian experimental design: A review}.
\newblock {\em Statistical Science}, 10(3):273--304, aug 1995.

\bibitem{cousinsSapsis2015_JFM}
W.~Cousins and T.~P. Sapsis.
\newblock {Reduced order precursors of rare events in unidirectional nonlinear
  water waves}.
\newblock {\em Journal of Fluid Mechanics}, 790:368--388, 2016.

\bibitem{fan19}
D.~Fan, G.~Jodin, T.~R. Consi, L.~Bonfiglio, Y.~Ma, L.~R. Keyes, G.~E.
  Karniadakis, and M.~S. Triantafyllou.
\newblock {A robotic Intelligent Towing Tank for learning complex
  fluid-structure dynamics}.
\newblock 5063:1--13, 2019.

\bibitem{Farazmande1701533}
M.~Farazmand and T.~P. Sapsis.
\newblock {A variational approach to probing extreme events in turbulent
  dynamical systems}.
\newblock {\em Science Advances}, 3(9):e1701533, 2017.

\bibitem{hill85abc}
P.~D. Hill.
\newblock {Kernel estimation of a distribution function}.
\newblock {\em Communications in Statistics - Theory and Methods}, 14:605--620,
  1985.

\bibitem{marzouk12}
X.~Huan and Y.~M. Marzouk.
\newblock {Gradient-based stochastic optimization methods in Bayesian
  experimental design}.
\newblock {\em International Journal for Uncertainty Quantification},
  4:479--510, dec 2014.

\bibitem{Kwong92}
M.~K. Kwong and A.~Zettl.
\newblock {\em {Norm Inequalities for Derivatives and Differences}}.
\newblock Springer Verlag Berlin, 1992.

\bibitem{Majda18ex}
A.~J. Majda, M.~N.~J. Moore, and D.~Qi.
\newblock {Statistical dynamical model to predict extreme events and anomalous
  features in shallow water waves with abrupt depth change}.
\newblock {\em Proceedings of the National Academy of Sciences},
  116:3982--3987, 2018.

\bibitem{Thomas10}
T.~Minka.
\newblock {Bayesian linear regression}.
\newblock 2010.

\bibitem{mohamad2016b}
M.~A. Mohamad, W.~Cousins, and T.~P. Sapsis.
\newblock {A probabilistic decomposition-synthesis method for the
  quantification of rare events due to internal instabilities}.
\newblock {\em Journal of Computational Physics}, 322:288--308, 2016.

\bibitem{mohamad2018}
M.~A. Mohamad and T.~P. Sapsis.
\newblock {Sequential sampling strategy for extreme event statistics in
  nonlinear dynamical systems}.
\newblock {\em Proc Natl Acad Sci U S A}, 115(44):11138--11143, 2018.

\bibitem{bilionis19}
P.~Pandita, I.~Bilionis, and J.~Panchal.
\newblock {Bayesian Optimal Design of Experiments For Inferring The Statistical
  Expectation Of A Black-Box Function}.
\newblock {\em arXiv}, jul 2018.

\bibitem{qi15}
D.~Qi and A.~J. Majda.
\newblock {Predicting Fat-Tailed Intermittent Probability Distributions in
  Passive Scalar Turbulence with Imperfect Models through Empirical Information
  Theory}.
\newblock {\em Submitted to Physica D}, 14:1687--1722, 2016.

\bibitem{rasmu05}
C.~E. Rasmussen and C.~K.~I. Williams.
\newblock {\em {Gaussian processes in machine learning}}.
\newblock The MIT Press, Cambridge, MA, 2005.

\bibitem{Rencher08}
A.~C. Rencher and B.~G. Schaalje.
\newblock {\em {Linear models in statistics}}.
\newblock John Wiley {\&} Sons, 2nd editio edition, 2008.

\bibitem{sapsis2018}
T.~P. Sapsis.
\newblock {New perspectives for the prediction and statistical quantification
  of extreme events in high-dimensional dynamical systems}.
\newblock {\em Phil. Trans. R. Soc. Lond. A}, 376:20170133, 2018.

\bibitem{Sarkar2018}
T.~Sarkar, M.~Roozbehani, and M.~A. Dahleh.
\newblock {Robustness Sensitivities in Large Networks}.
\newblock In {\em Emerging Applications of Control and Systems Theory}, pages
  81--92. 2018.

\bibitem{Ortiz05}
S.~Serebrinksy and M.~Ortiz.
\newblock {A hysteretic cohesive-law model of fatigue-crack nucleation}.
\newblock {\em Scripta Materialia}, 53:1193--1196, 2005.

\bibitem{Verdinelli1992}
I.~Verdinelli and J.~B. Kadane.
\newblock {Bayesian designs for maximizing information and outcome}.
\newblock {\em Journal of the American Statistical Association},
  87(418):510--515, 1992.

\end{thebibliography}

\pagebreak

\section*{Appendix A: Gradient of trace criteria}\label{trace_cr_gr}
Several criteria in this work take the form\begin{equation}
\lambda[\mathbf{h}]=\text{tr}[\mathbf{S}_{xx}'^{-1} \mathbf{C}], 
\end{equation}
where $\mathbf{C}$ is a symmetric matrix and $\mathbf{S}_{xx}'=\mathbf{S}_{xx}+\mathbf{hh}^T.$ The gradient of this expression can be explicitly computed. We first note that
\begin{equation*}
\frac{\partial \mathbf{S'}_{xx}^{-1}}{\partial h_k}=\mathbf{-S'}_{xx}^{-1}\frac{\partial
(\mathbf{h} \mathbf{h}^T) }{\partial h_k}\mathbf{S'}_{xx}^{-1},
\end{equation*}
where,
\begin{equation*}
\frac{\partial
(\mathbf{h} \mathbf{h}^T) }{\partial h_k}=\delta_{ik}h_j+\delta_{kj}h_i.
\end{equation*}
In this way we will have
\begin{align*}
\frac{\partial \lambda}{\partial h_k}& =-\text{tr}[\mathbf{S}_{xx}'^{-1}\frac{\partial
(\mathbf{h} \mathbf{h}^T) }{\partial h_k}\mathbf{S}_{xx}'^{-1}\mathbf{C}]\\
& =-[\mathbf{S}_{xx}'^{-1}]_{ij}(\delta_{jk}h_m+\delta_{km}h_j)[\mathbf{S}_{xx}'^{-1}]_{mn}[\mathbf{C}]_{ni}\\
& =-h_m[\mathbf{S}_{xx}'^{-1}]_{mn}[\mathbf{C}]_{ni}[\mathbf{S}_{xx}'^{-1}]_{ik}-[\mathbf{S}_{xx}'^{-1}]_{kn}[\mathbf{C}]_{ni}[\mathbf{S}_{xx}'^{-1}]_{ij}h_j\\
& =-\mathbf{h}^T\mathbf{S}_{xx}'^{-1}\mathbf{C}\mathbf{S}_{xx}'^{-1}-(\mathbf{S}_{xx}'^{-1}\mathbf{C}\mathbf{S}_{xx}'^{-1}\mathbf{h})^T\\
& =-2\mathbf{h}^T\mathbf{S}_{xx}'^{-1}\mathbf{C}\mathbf{S}_{xx}'^{-1}.
\end{align*}
For the case of nonlinear regression $\mathbf{h}=\phi(\mathbf{z})$. Then\begin{equation*}
\frac{\partial
(\phi(\mathbf{z}) \phi(\mathbf{z})^T) }{\partial z_k}=\frac{\partial \phi_i}{\partial z_k}\phi_j+\frac{\partial \phi_j}{\partial
z_k}\phi_i.
\end{equation*}
In this way we will have
\begin{align}
\begin{split}
\frac{\partial \lambda}{\partial z_k}& =-[\mathbf{S}_{\phi\phi}'^{-1}]_{ij}(\frac{\partial \phi_j}{\partial z_k}\phi_m+\frac{\partial \phi_m}{\partial
z_k}\phi_j)[\mathbf{S}_{\phi\phi}'^{-1}]_{mn}[\mathbf{C}]_{ni}\\
& =-2[\phi^T\mathbf{S}_{\phi\phi}'^{-1}\mathbf{C}\mathbf{S}_{\phi\phi}'^{-1}]_{j}\frac{\partial
\phi_j}{\partial z_k}.
\end{split}
\end{align}
\clearpage
\section*{Appendix B: Optimal approximation of $p_y^{-1}$}\label{trace_cr_gr}
To approximate the inverse of the output pdf, $\frac{1}{p_y}$ over $S_y$ we are going to use a least square approach. Specifically, we will assume a symmetric output density and we will employ the approximation\begin{equation}
\frac{1}{p_y(y)}\simeq\frac{1}{p_y(0)}+p_{2} (y-\mu_y)^2.
\end{equation}
The constant $p_{2}$ is chosen so that we have optimal least square approximation over the interval $[\mu_y,\mu_y+\beta \sigma_y]$ where $\beta$ is a fixed parameter that defines the output region of interest, $S_y$. By direct minimization we obtain\begin{equation*}
p_{2}=\frac{5}{\beta^5\sigma_y^3}\int_{\mu_y}^{\mu_y+\beta \sigma_y}\frac{y^{2}}{p_y(y)}dy-\frac{5}{3\beta^{2}p_y(\mu_y)}.
\end{equation*}
For the case of Gaussian output the above expression takes the form\begin{equation*}
p_{2}=\frac{5 \sqrt{2\pi}}{\beta^5\sigma_y}\left(\int_{0}^{\beta}{z^{2}}e^{\frac{z^2}{2}}dz-\frac{\beta^{3}}{3} \right).
\end{equation*}
In this way the least square approximation over the interval $[\mu_y,\mu_y+\beta \sigma_y]$ will be \begin{equation}
\frac{1}{p_y(y)}\simeq \sqrt{2\pi}\sigma_y+\frac{5 \sqrt{2\pi}}{\beta^5\sigma_y}\left(\int_{0}^{\beta}{z^{2}}e^{\frac{z^2}{2}}dz-\frac{\beta^{3}}{3}
\right) (y-\mu_y)^2.
\end{equation}
We denote the criterion with the coefficients obtained from this approximation as $Q_{\beta}$. Specifically, 
\begin{align}
\begin{split}
Q_{\beta \sigma}(\mathbf{h})\frac{1}{\sigma_V^2}& = \sqrt{2\pi}\sigma_{y_0}(1+\text{tr}[\mathbf{S}'^{-1}_{xx}\mathbf{C}_{xx}]+\mathbf{\mu}_{x}^T\mathbf{S}'^{-1}_{xx}{\mu_x})\\ & + \frac{5 \sqrt{2\pi}}{\beta^5\sigma_{y_0}}\left(\int_{0}^{\beta}{z^{2}}e^{\frac{z^2}{2}}dz-\frac{\beta^{3}}{3}
\right) \left( c_{0}(1+{\mu}_{x}^T\mathbf{S}'^{-1}_{xx}{\mu_x})
 +2\text{tr}[\mathbf{S}_{xx}^{-1}\mathbf{S}^T_{yx}\mathbf{}\mathbf{S}_{yx}\mathbf{S}_{xx}^{-1}\mathbf{C}_{xx}\mathbf{S}'^{-1}_{xx}\mathbf{C}_{xx}] \right).
\end{split}
\end{align}
For large values of $\beta$ we have $\lim_{\beta \rightarrow \infty}\kappa=\infty$ and the $Q$ criterion is essentially dominated by the output-dependent term. For the case of very small $\beta$ we have $\lim_{\beta \rightarrow 0}\kappa=\frac{\sqrt{2\pi}}{\sigma_y}$.
\clearpage
\section*{Appendix C: Effect of the weights in the $Q$ criterion}
\begin{figure}[h]
\begin{center}
\includegraphics[width=0.68\textwidth]{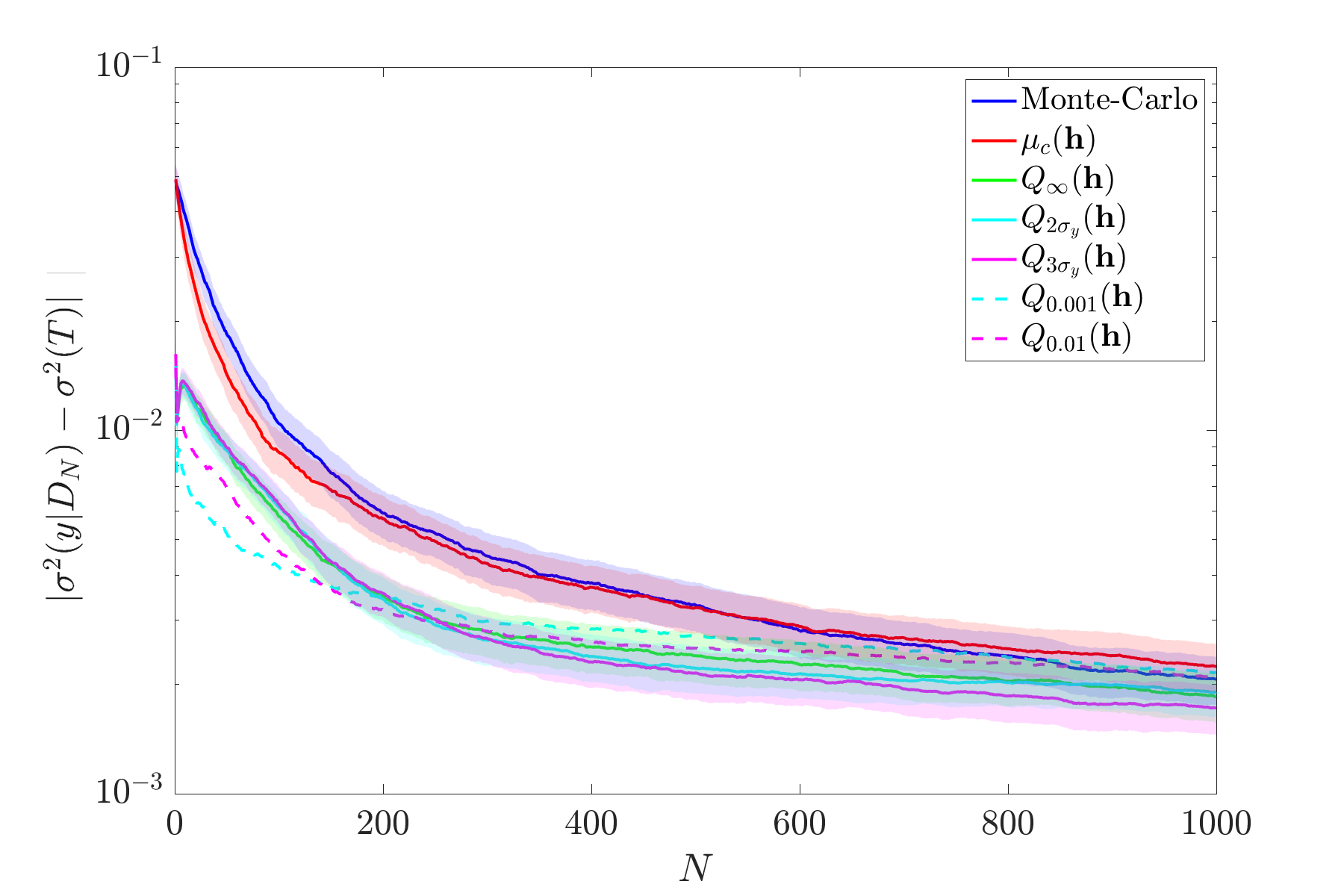}
\end{center}
\setlength{\abovecaptionskip}{-0pt}
\caption{More detailed results for Case I of the high dimensional problem.
The effect of the $\beta$ parameter is shown. While for the first iterations
it plays no role, asymptotically it improves the behavior of the sampling
scheme.}
\label{fig_highD_curves_more}
\end{figure}

\begin{figure}[t]
\begin{center}
\includegraphics[width=0.98\textwidth]{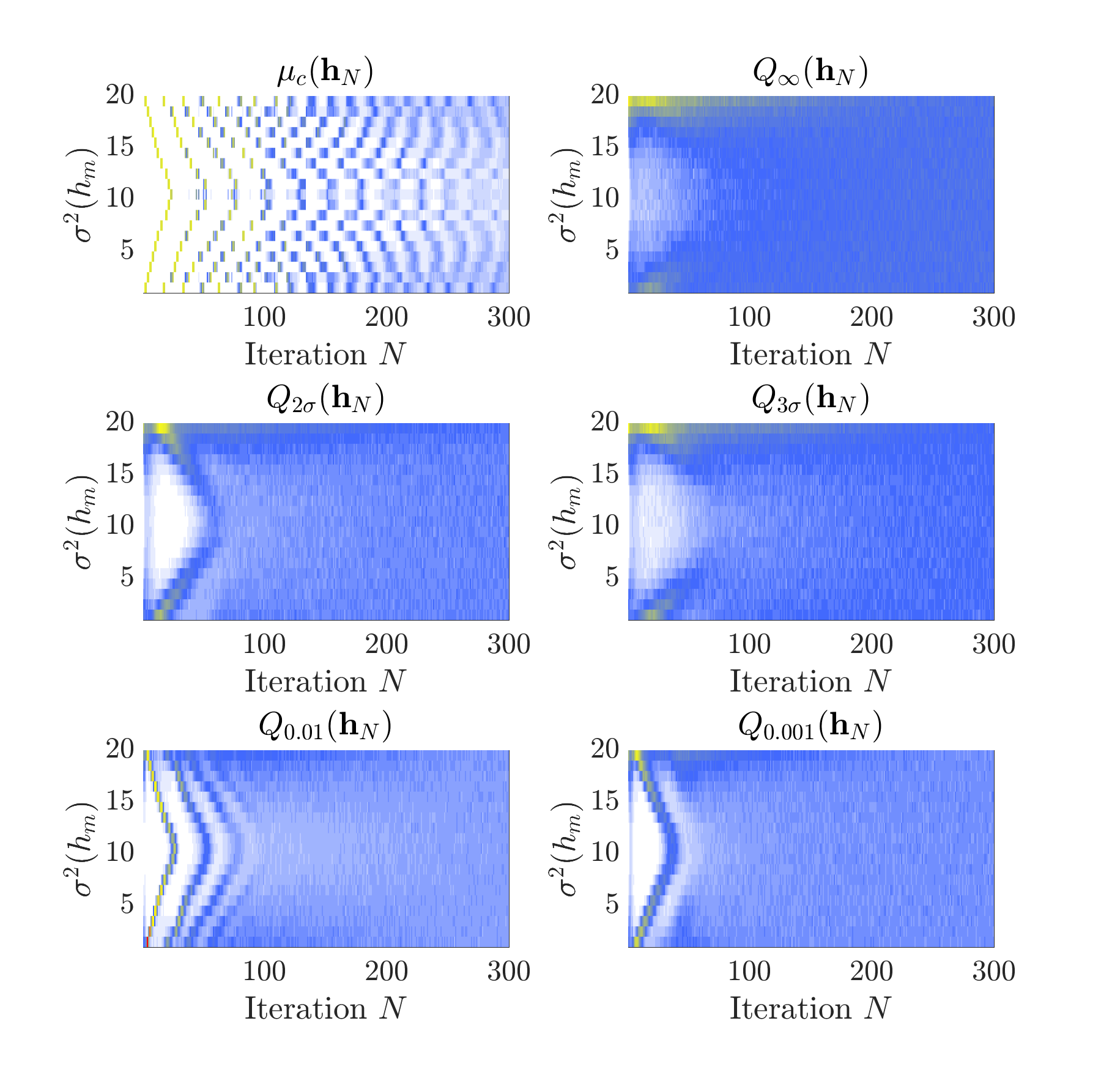}
\end{center}
\setlength{\abovecaptionskip}{-0pt}
\caption{More detailed results for the samples of $\mathbf{h}$ with respect
to the number of iteration $N$ for Case I of the high dimensional problem.  The effect of the $\beta$ parameter is shown.}
\label{fig_highD_samples_more}
\end{figure}

Here we present additional results for Case I of the high dimensional system.  Specifically, in Figure $\ref{fig_highD_curves_more}$ we present the performance of the sampling algorithm according to various choices of the $\beta$ parameter.
The corresponding sampling patterns are shown in (Figure \ref{fig_highD_samples_more}). All cases presented are averaged over $L=500$ numerical experiments to remove the effect of observational noise.

We observe that for $\beta=2$ or $\beta=3$ the performance for small $N$ is very close to the one obtained with $Q_{\infty}$. For larger $N$, however, the performance with finite $\beta$ is improved. In addition to the finite $\beta$ case, we also present two cases with fixed $p_{1}, p_2$. Specifically, we have $Q_{0.01}$ representing the case $p_1=0.01$ and $p_2=1$, while  $Q_{0.001}$ represents the case $p_1=0.001$
and $p_2=1$. It is interesting to observe that   $Q_{0.001}$  has better performance for small $N$ compared with all other criteria.

\clearpage
\section*{Appendix D: The case of unknown $\sigma_V^2$}

Here we consider the case of a priori unknown covariance $\sigma_V^2$. To simplify
the presentation we will restrict our analysis to scalar output and vector
input. We formulate a linear regression model with an input vector $\mathbf{x}$
that multiplies a coefficient vector $\mathbf{a}$ to produce an output scalar
${y}$, with Gaussian noise added that has unknown variance:
\begin{align}\label{linear_mod}
\begin{split}
{y}\ & = \mathbf{a}^T\mathbf{x}+{e},\\
{e} & \sim\ \mathcal{N}(0,\sigma_V^2), \\
p(y|\mathbf{x},\mathbf{a},\sigma^2) & = \mathcal{N}(\mathbf{a}^T\mathbf{x},\sigma^2).
\end{split}
\end{align}
For the vector $\mathbf{a}$ we assume a Gaussian prior with mean $\mathbf{m=0}$
and  covariance $\mathbf{K=I}\alpha$, where $\alpha$ is a parameter that
will be optimized with respect to the evidence. This has the
form:\begin{align}
p(\mathbf{a}) & \sim\mathcal{N}(0,\mathbf{I}\alpha) .
\end{align}
A conjugate prior for $\sigma_V^2$ is the inverse Gamma (or inverse Wishart
in the multi-dimensional case):\begin{equation}
p(\sigma_V^2)=\frac{q(\sigma^{2}_0,\nu)}{(\sigma_V^2)^{1+\frac{\nu}{2}}}\exp\left(-\frac{\sigma_0^2}{2\sigma_V^2}\right),
\end{equation}
where $q$ is a normalization constant: $q=\frac{\sigma_0^{\nu}}{2^{\frac{\nu}{2}}\Gamma(\nu/2)}$,
$\sigma_0^2$ is a prior value for $\sigma_V^2$ and $\nu$ is a parameter that
is optimized via empirical Bayes. 

The posterior for the unknown coefficients will be given by eq. (\ref{coef_dis}), while the posterior for $\sigma_V^2$ takes the form\begin{equation}\label{var_post}
p(\sigma_V^2|D)=\frac{q(\sigma_{\mathbf{Y}|\mathbf{X}}^2+\sigma^{2}_0,N+\nu)}{(\sigma_V^2)^{1+\frac{N+\nu}{2}}}\exp\left(-\frac{\sigma_{\mathbf{Y}|\mathbf{X}}^2+\sigma^{2}_0}{2\sigma_V^2}\right),
\end{equation}
where, $\sigma_{\mathbf{Y}|\mathbf{X}}^2= \mathbf{Y}\mathbf{Y}^T-(\alpha+1)^{{-1}}\mathbf{S}_{yx}\mathbf{S}_{{xx}}^{-1}\mathbf{S}^T_{yx}$.
Multiplying the predictive distribution (\ref{model1}) with the posterior
for $\sigma^2$ and integrating over this argument we have the predictive
pdf ($t-$distributed):
\begin{equation}
\label{pred_pdft}
p(\mathbf{y}|\mathbf{x},D) = \mathcal{T}(\mathbf{S}_{yx}\mathbf{S}_{xx}^{-1}\mathbf{x},(\sigma_{\mathbf{Y}|\mathbf{X}}^2+\sigma^2_0)(1+c)^{-1},N+\nu+1)
\end{equation}
where the parameters $(\alpha, \nu)$ are chosen by maximizing the evidence
(set the gradient of $p(\mathbf{Y}|\mathbf{X})$ equal to zero), which leads
to the following fixed-point problem:
\begin{align}
\begin{split}
\hat \sigma^2 & = \frac{\sigma_{\mathbf{Y}|\mathbf{X}}^2+\nu}{N+\nu}, \\
\alpha&=\frac{m}{ (\hat \sigma^2)^{-1}\mathbf{S}_{y{x}}\mathbf{S}_{xx}\mathbf{S}^T_{y{x}}-m},\\
\nu^{new}&=\nu\frac{\Psi(\frac{N+\nu}{2})-\Psi(\frac{\nu}{2})}{\log\left(\frac{\sigma_{\mathbf{Y}|\mathbf{X}}^2}{\nu}+1
\right)+(\hat \sigma^2)^{-1}-1},
\end{split}
\end{align}
where $\Psi(x)=\frac{d\log{\Gamma{(x)}}}{dx}$ is the digamma function.

\subsection*{Selecting inputs by maximization of the mutual information}
We follow the same steps as in the known variance case. We hypothesize
a new sample, $\mathbf{x}_{N+1}=\mathbf{h} \in \mathbb{S}^{m-1}$ and the
goal
is maximizing the entropy transfer or mutual information between the input
and output variables, when this new sample is added. For this case of a priori unknown variance $\sigma_{V}^2$ we denote the mutual information as $\mathcal{\hat I}$.\ We will have \begin{equation}
\mathcal{\hat I}(\mathbf{x,y}|D') =\mathcal{E}_{x}+\mathcal{E}_{y|D'}-\mathcal{E}_{x,y|D'}.
\end{equation}  
Following the same steps with section \ref{entr_conv} we have for the entropy    of $p(\mathbf{x,y}|D')$:\begin{align*}
\mathcal{E}_{x,y}(\mathbf{h}) & = \mathbb{E}^{x}[\mathcal{E}_{y|x}(\mathbf{x;h})]+\mathcal{E}_{x}.
\end{align*}
We focus on computing the first term on the right hand side. In this case
of a priori unknown variance the conditional output follows the $t-$student
distribution (eq. (\ref{pred_pdft})). Using standard expressions for its
entropy we have (setting $N'=\nu+N+1$) 
\begin{align*}
\mathcal{E}_{y|x}(\mathbf{x;h})=\frac{N'+1}{2}\left(\Psi \left(\frac{N'+1}{2}\right)-\Psi\left(\frac{N'}{2}\right)\right)+\log\left(\sqrt{N'}B\left(\frac{N'}{2},\frac{1}{2}\right)\right)
\\-\frac{1}{2}\log
(1+c(\mathbf{x;h}))+\frac{1}{2}\log(\sigma_{\mathbf{Y}'|\mathbf{X}'}^2+\sigma^2_0),
\end{align*}
where,
\begin{align}
\label{sigma_c}
\begin{split}
\sigma^2_{\mathbf{Y}'|\mathbf{X}'}& =\mathbf{Y'}\mathbf{Y'}^T-(\alpha+1)^{{-1}}\mathbf{S}_{yx}'\mathbf{S}_{{xx}}'^{-1}\mathbf{S}'^T_{yx}\\
& =\mathbf{Y}\mathbf{Y}^T+\mathbf{(S}_{yx}\mathbf{S}_{{xx}}^{-1}\mathbf{h})^{2}-(\alpha+1)^{{-1}}\mathbf{S}_{yx}\mathbf{S}_{{xx}}^{-1}[\mathbf{S}_{yx}+\mathbf{S}_{yx}\mathbf{S}_{xx}^{-1}\mathbf{hh}^T]^{T}\\
& =\sigma^2_{\mathbf{Y}|\mathbf{X}}+\frac{\alpha}{1+\alpha}\mathbf{(S}_{yx}\mathbf{S}_{{xx}}^{-1}\mathbf{h})^{2}.
\end{split}
\end{align} 
Note that in the second equality we used eq. (\ref{mean_inv}).
 In  general, we cannot compute analytically the entropy of the output, conditional
on $D'$. To this end, the mutual information of the input and output, conditioned
on $D',$ takes the form\begin{equation}\label{info_full}
\mathcal{\hat I}(\mathbf{x,y}|D') =\mathcal{E}_{y}(\mathbf{h})-\frac{1}{2}\mathbb{E}^{x}[\log(1+c(\mathbf{x;h}))]+\frac{1}{2}\log(\sigma_{\mathbf{Y'}|\mathbf{X'}}^2(\mathbf{h})+\sigma^2_0)+R,
\end{equation}
where $R$ are terms that do not depend on the new point $\mathbf{h}$.
The second and third terms are computed for each $\mathbf{h}$ with direct Monte-Carlo using $10^5$ samples. It is important to emphasize that the mutual information criterion with unknown variance, $(\ref{info_full})$, depends on the output values $\mathbf{Y }$ (through the third term on the right hand side), in contrary to the known variance case  (section \ref{entr_conv}). However, as it can be seen from eq. (\ref{sigma_c}) this dependence can be very weak or even zero depending on the value of the parameter $\alpha$ which is chosen based on maximization of the evidence.

In the last expression the term with the highest computational cost is the entropy of the output (first term) as one needs to estimate the histogram of $y$. An analytical approximation can be obtained based on a Gaussian assumption for the output, $y$. In this case the mutual information takes the form: \begin{equation}\label{info_full1}
\mathcal{\hat I}_G(\mathbf{x,y}|D') =\frac{1}{2}\log(2\pi e \sigma_y^2(\mathbf{h}))-\frac{1}{2}\mathbb{E}^{x}[\log(1+c(\mathbf{x;h}))]+\frac{1}{2}\log(\sigma_{\mathbf{Y'}|\mathbf{X'}}^2(\mathbf{h})+\sigma^2_0)+R,
\end{equation}
where $ \sigma_y^2(\mathbf{h})$ is the estimated variance of the output after a new candidate input $\mathbf{h}$. This is computed with Monte-Carlo simulation of $y$, i.e. generate $10^5$ random realizations using (\ref{var_post}), (\ref{coef_dis})
and (\ref{linear_mod}).
\subsection*{$Q-$criterion with unknown output variance}

We emphasize that the selection approach using the $Q$ criterion is not modified at all for the case of unknown output variance. This is because the unknown variance $\sigma_V^2$  appears as a multiplication factor in the $Q$ criterion (eq. (\ref{Q_com})), i.e. $\sigma_V^2$ is re-estimated each time a new data point is added using equation (\ref{Q_com}) but its value does not modify the optimal sample $\mathbf{h}$.
\subsection*{Numerical comparison for the 2d linear problem}
\begin{figure}[ht]
\begin{center}
\includegraphics[width=0.95\textwidth]{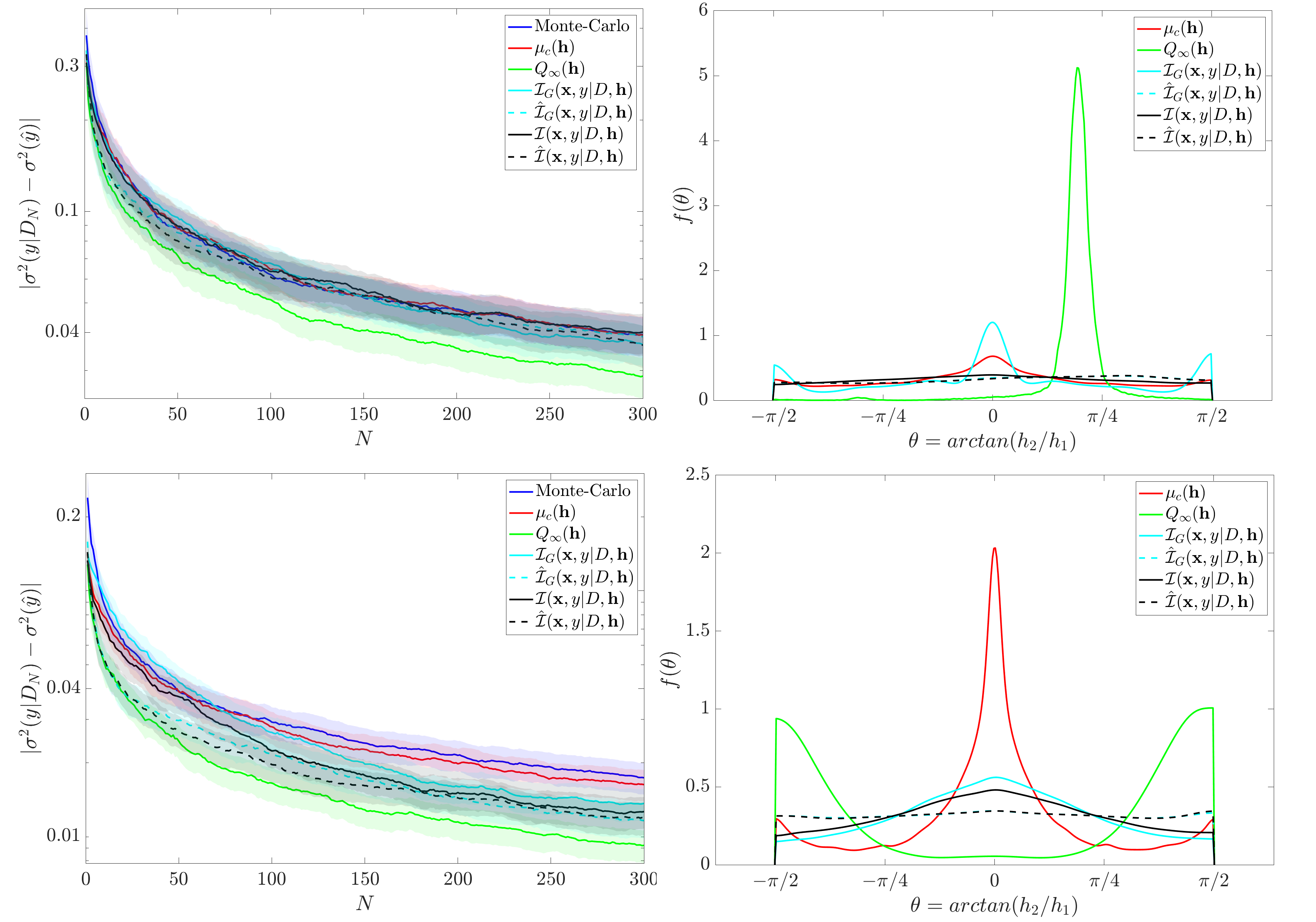}
\end{center}
\caption{Comparison of selection methods based on different criteria
and the Monte-Carlo method including also the case of unknown $\sigma_V^2$. Problem setup and parameters are as in section \ref{linear2d} and Figure $\ref{fig:2d}$. The following methods are shown: mean square error, $\mu_c$; $Q-$criterion; Gaussian approximation of mutual information with known variance, $\mathcal{ I}_G$, and with
unknown variance, $\mathcal{\hat I}_G$; directly computed mutual information
with 
known variance, $\mathcal{ I}$, and unknown variance, $\mathcal{\hat I}$. The $Q-$criterion does not depend on whether the noise variance is known or estimated. Note that the pdf of samples (right plots) between $\mathcal{\hat I}$ and $\mathcal{\hat I}_G$ overlap, i.e. are indistinguishable.}
\label{fig:appD}
\end{figure}

Results and direct comparison with the case of known $\mathbf \sigma_V^2$ are shown
in Figure \ref{fig:appD} for the linear problem of section \ref{linear2d} with a two-dimensional input. For all methods shown we have run $400$ experiments (as we did for Fig. $\ref{fig:2d}$). In the higher dimensional problem the approach based on direct computation of mutual information is not applicable due to the vast computational cost.  As we can observe the selection process based on mutual information with unknown output variance ($\mathcal{\hat I}_G$ or $\mathcal{\hat
I}$), has slightly improved performance compared with the case of known $\sigma_V^2$ ($\mathcal{
I}_G$ or $\mathcal{
I}$) but this improvement is observed only for very small number of samples. As $N$ increases the criterion with unknown output variance is comparable with mutual information with given output variance ($\mathcal{
I}_G$ or $\mathcal{
I}$). This is because the optimal value of $\alpha$ tends to zero as $N$ increases and therefore the criteria with known and unknown output variance become practically identical.

It is important to emphasize that beyond the faster convergence of the $Q-$criterion, its main advantage is the low computational cost (many orders of magnitude  smaller compared with the methods based on mutual information either with direct computation or through a Gaussian approximation) which allows applicability to higher dimensional problems.  

\end{document}